\documentclass[a4paper,10pt,twoside]{article}

\usepackage{clin}        
\usepackage{harvard}     
\usepackage{appendix}
\usepackage{hyperref}
\usepackage{pdflscape}
\usepackage{makecell}

\usepackage{graphicx}
\usepackage{subcaption}
\usepackage{listings}
\usepackage{xcolor}

\definecolor{bgcolor}{rgb}{0.95, 0.95, 0.95}       
\definecolor{keywordcolor}{rgb}{0.0, 0.25, 0.75}   
\definecolor{stringcolor}{rgb}{0.75, 0.0, 0.25}    
\definecolor{commentcolor}{rgb}{0.4, 0.5, 0.4}     
\definecolor{framecolor}{rgb}{0.8, 0.8, 0.8}       

\lstdefinestyle{modern}{
    backgroundcolor=\color{bgcolor},
    basicstyle=\ttfamily\small,
    keywordstyle=\color{keywordcolor}\bfseries,
    stringstyle=\color{stringcolor},
    commentstyle=\color{commentcolor}\itshape,
    showstringspaces=false,
    showspaces=false,
    showtabs=false,
    frame=single,
    framerule=0.5pt,
    rulecolor=\color{framecolor},
    breaklines=true,
    breakatwhitespace=true,
    columns=fullflexible
}

\lstdefinestyle{python}{
    style=modern,
    language=Python,
    morekeywords={def, return, if, else, for, while, import, as, class, from, with, try, except, finally, lambda, yield}
}

\lstdefinelanguage{yaml}{
    style=modern,
    keywords={true, false, null, Yes, No, On, Off},
    keywordstyle=\color{keywordcolor}\bfseries,
    comment=[l]{\#},
    morestring=[b]',
    morestring=[b]",
    moredelim=[l][\color{black}]{\&},
    moredelim=[l][\color{black}]{*},
    sensitive=true
}

\lstdefinelanguage{jinja}{
    style=modern,
    keywords={for, endfor, if, endif, block, endblock, set, endset, macro, endmacro, import, from},
    keywordstyle=\color{keywordcolor}\bfseries,
    comment=[l]{\#},
    morestring=[b]',
    morestring=[b]",
    sensitive=true
}

\begin{document}

\title{Fietje: An open, efficient LLM for Dutch}

\author{Bram Vanroy$^{1,2}$ \email{bram.vanroy@kuleuven.be}
\AND \addr{$^1$KU Leuven, Blijde Inkomststraat 21, 3000 Leuven, Belgium}
\AND \addr{$^2$Dutch Language Institute, Rapenburg 61, 2311 GJ Leiden, The Netherlands} }

\maketitle\thispagestyle{empty} 


\begin{abstract}This paper introduces Fietje, a family of small language models (SLMs) specifically designed for the Dutch language. The model is based on Phi 2, an English-centric model of 2.7 billion parameters. Fietje demonstrated competitive results with larger language models upon its release. A core emphasis of this work is transparency and reproducibility: Fietje is fully open-source, with model weights, datasets, training, and evaluation code all publicly accessible.

The paper discusses the performance of Fietje and many other models on an extensive evaluation suite of benchmarks on reasoning, sentiment analysis, world knowledge, linguistic acceptability and word sense disambiguation. Evaluation results illustrate the rapid progress in the field of LLMs, where recent small models outperform older, larger models that were fine-tuned for Dutch. This trend signals an exciting future for Dutch language processing, suggesting that even compact LLMs are becoming increasingly capable. Furthermore, ongoing and future efforts to adapt LLMs to Dutch are poised to enhance these models even further, broadening their applicability and accessibility. Fietje is only an intermediate step in improving accessibility to language technology for users of the Dutch language.
\end{abstract}

\section{Introduction}

Large Language Models (LLMs) have revolutionized natural language processing, demonstrating remarkable proficiency across various tasks, including generation tasks as well as zero-shot classification or annotation. However, their performance has predominantly focused on English, leaving other languages underrepresented. For this reason, we introduce Fietje, a continued pretraining model that adapts the predominantly English-focused phi-2 \cite{javaheripi2023phi2} Small Language Model (SLM) of 2.7 billion parameters to Dutch alongside instruction and chat variants. By training on 28 billion Dutch tokens sourced from open, filtered web data, Fietje enhances its proficiency in the Dutch language.

Since its initial release in April 2024, many new LLMs have been published. More and more, LLMs are moving away from English-only data and have started incorporating multilingual capabilities, such as the Phi 3.5 \cite{abdin2024phi3}, Qwen 2.5 \cite{qwen2.5} and Llama 3.1, 3.2 and 3.3 \cite{grattafiori2024llama3} models.

In addition to describing the creation of Fietje and the datasets used, it is evaluated on key benchmarks in reasoning, knowledge, sentiment analysis, linguistic acceptability, and word sense disambiguation. For comparison, other models are also evaluated. These models cover different sizes (2-7B parameters) as well as English-centric, multilingual, and Dutch-adapted models. The benchmark results reveal that at the time of publication Fietje surpassed expectations for its size and showed performance that is at times competitive with models twice its size. However, at the same time the benchmarks also illustrate the rapidly changing landscape of LLMs, and how small, multilingual models are now outperforming even Dutch-specific models. At the end, a reflective discussion section posits some points of focus for future models for Dutch, and a limitation section summarizes shortcomings of Fietje and its evaluation.

By emphasizing transparency and openness through the release of data, training and evaluation code, and model weights, this work aims to advance the development and assessment of efficient, high-performing LLMs and SLMs for underrepresented languages like Dutch. Training code and evaluation be accessed via Github on \url{https://github.com/BramVanroy/fietje-2} and \url{https://github.com/BramVanroy/clin34-benchmarks} respectively. Models, including quantized versions, and data are available on the Hugging Face hub \url{https://huggingface.co/collections/BramVanroy/fietje-2-662cb803ed5cc4f617404146}.

\section{Related work}\label{sec:related-work}

Adapting large language models (LLMs) to new languages has garnered significant attention, especially as models trained primarily on English often exhibit performance limitations in other languages. To improve access to language technology to other languages, researches have looked into adapting existing English-focused models to other languages. Strategies for language adaptation generally fall into three categories: tokenizer updates, partial model retraining, and input-level modifications.

\subsection{Tokenizer updates}

\citeasnoun{remy2024transtokenization} propose a cross-lingual vocabulary transfer strategy called trans-tokenization. This method initializes token embeddings in the target language using a weighted average of semantically similar tokens from the source language. By leveraging a parallel corpus, trans-tokenization achieves efficient vocabulary adaptation, enhancing performance in low-resource scenarios without extensive retraining. Similarly, \citeasnoun{csaki-etal-2024-sambalingo} demonstrate that augmenting the tokenizer vocabulary with language-specific tokens can reduce the number of subwords needed to encode words in the target language. However, while these methods improve tokenizer efficiency, the performance gains on downstream tasks remain inconsistent. The Dutch Tweety model proposed in \citeasnoun{remy2024transtokenization} is part of the benchmark suite in this paper.

\subsection{Model retraining}

Another efficient strategy involves retraining the embedding layer while keeping the transformer layers frozen. \citeasnoun{de-vries-nissim-2021-good} adapted GPT-2 for Dutch and Italian by retraining the lexical embeddings. This approach minimizes computational costs and retains the original model's knowledge. Nevertheless, the method can still result in syntactic and lexical errors when significant linguistic divergence exists between the source and target languages. 

Continued pretraining on target-language corpora is a powerful approach to adapt LLMs to new languages. \citeasnoun{toraman-2024-adapting} applied continual pretraining to English-centric models for Turkish, observing significant gains in language comprehension and downstream performance. However, interestingly, for Turkish they note that vocabulary extension alone offers negligible benefits. For Dutch, \citeasnoun{vanroy2023languageresourcesdutchlarge} attempted to retrain Llama 2 but the real performance improvements in the Dutch LLM space only broke through with the release of GEITje, a continued-pretrained version of Mistral 7B v0.1 \cite{jiang2023mistral7b} for Dutch \cite{rijgersberg2023geitje}. Later, GEITje was further improved through preference alignment \cite[GEITje 7B Ultra]{vanroy2024geitje7bultraconversational}. These models outperformed previous adaptations due to their extensive use of Dutch-specific data. 

Continue-pretraining is expensive as it often involves training on large corpora. Techniques like QLoRA \cite{dettmers2023qlora} reduce computational requirements by fine-tuning only a limited number of parameters. However, \citeasnoun{vanroy2023languageresourcesdutchlarge} and \citeasnoun{toraman-2024-adapting} highlight that QLoRA, while efficient, may not achieve the same quality as full retraining, particularly when adapting to languages with complex morphology.

\subsection{In-context learning and prompt engineering}

Rather than retraining any of the parameters of a model, in-context learning offers a way to adapt LLMs to new languages by prompting. \citeasnoun{zhang-etal-2024-teaching} introduce a framework, which adapts LLMs to new languages on the fly by providing dictionaries and a small number of parallel examples in the prompt. Their approach enables translation and other tasks without any model parameter updates, achieving significant performance gains for unseen low-resource languages.

Similarly, cross-lingual prompting methods have also been explored. \citeasnoun{huang-etal-2023-languages} propose cross-lingual ``thought prompting'', where models are guided through language-specific prompts to enhance multilingual reasoning, intermediately translating queries to a different language. This is similar to the work of \citeasnoun{qin-etal-2023-cross}, who show that translating prompts into a high-resource language (e.g., English) and applying chain-of-thought reasoning can improve performance on multilingual tasks. These approaches leverage the model's stronger performance in languages it was originally trained on, circumventing the need for retraining.

\subsection{Hybrid methods}

Combining multiple adaptation strategies often yields the best results. For instance, \citeasnoun{remy2024transtokenization} and \citeasnoun{toraman-2024-adapting} suggest that combining tokenizer updates with continued pretraining can mitigate the weaknesses of either method used in isolation. Similarly, \citeasnoun{joshi2024adaptingmultilingualllmslowresource} advocate for integrating efficient fine-tuning techniques like LoRA with careful prompt engineering to maximize performance in low-resource scenarios while at the same time minimizing the need for large quantities of computational power.

Future research directions may focus on developing more efficient cross-lingual transfer techniques and improving the scalability of hybrid adaptation strategies. Some insights gathered from this paper and related work will be discussed in the Discussion.

\section{Model creation}\label{sec:creation}

Three different flavors of Fietje were created: a base model for text completion, an instruct version for instruction following, and a chat version for an improved assistant experience. All models were trained on hardware provided by the Flemish Supercomputer Center.\footnote{\url{https://www.vscentrum.be/}} As for training code, the Alignment Handbook \cite{tunstall2024alignment} Github repository already allowed for instruction tuning (supervised funetuning or SFT) and preference tuning. I updated the code to enable continued pretraining and merged our changes to that repository so that other users can also continue-pretrain their models with this library. All the model training is fully reproducible thanks to the open data and open code. The necessary configuration file and training instructions can be found on Github: \url{https://github.com/BramVanroy/fietje-2/tree/main/training}.

\subsection{Base}

\paragraph{Architecture} Fietje's base model is built on the foundation of Phi 2 \cite{javaheripi2023phi2}\footnote{\url{https://huggingface.co/microsoft/phi-2}}, a Transformer \cite{vaswani2017attention} decoder model with a context length of 2048 tokens. It is a base model with no explicit chat or instruction tuning, although the model README file suggests three formats that work best for question-and-answering, chat between fictional characters, and coding. Phi 2 was released in the middle of December 2023 as a small language model of 2.78 billion parameters that achieved performance on the level of much larger models. It was trained with an emphasis on the dataset of 250 billion tokens, totalling 1.4 trillion tokens after over-sampling. Microsoft's approach for Phi 2 was built on their early work with Phi 1 \cite{gunasekar2023textbooksneed} and Phi 1.5 \cite{li2023textbooksneediiphi15}, where they highlighted the importance of high-quality, curated data rather than overly relying on massive-scale, but potentially poor quality, web data. In that work, aptly titled ``Textbooks are all you need'', they showed remarkable performances on benchmarks, not by increasing parameter count or simply scaling compute and data, but by filtering existing code or web data on its quality, and by generating synthetic data of diverse topics. Unfortunately Phi 2 is an English-centric model. Multilingual Phi models only saw the light of day starting with Phi 3.5, released in August 2024, long after Fietje's release. Therefore, Fietje set out to adapt Phi 2 to Dutch. Phi 2 was selected as the starting point because at the time it was the best performing small language model (less than 3 billion parameters).

\paragraph{Data} To adapt Phi 2 to Dutch, it was continue-pretrained on 28 billion high-quality Dutch tokens. For comparison: GEITje 7B, a larger model continue-pretrained for Dutch based on Mistral was trained on 10 billion tokens \cite{rijgersberg2023geitje}, similar to Boreas 7B,\footnote{\url{https://huggingface.co/yhavinga/Boreas-7B}} another Mistral 7B \cite{jiang2023mistral7b} finetune. So even though Phi 2 is a much smaller model, it was continue-pretrained on much more Dutch specific data to ensure a high-quality, small model. The training dataset consists of Dutch Wikipedia (dump of November 2023\footnote{\url{https://huggingface.co/datasets/wikimedia/wikipedia}}) and was further expanded with random samples from the CulturaX dataset \cite{nguyen-etal-2024-culturax}. CulturaX is a high-quality, multilingual dataset covering 167 languages. Its creation is thorough, including URL-based filtering as well as metric-based cleaning (e.g., perplexity score, characters per document, number of lines) and deduplication. Before training, Dutch CulturaX was filtered even further (Wikipedia was not filtered in this way, as discussed later):

\begin{itemize}
  \itemsep-0.5em
  \item removed documents that contain the text ``rechten voorbehouden'' or ``rights reserved'' to avoid including explicitly copyrighted materials;
  \item remove documents whose URL contained ``wikipedia.org'' (because a very clean version of Wikipedia was included separately);
  \item removed documents that contain a ``bad word'' (see Appendix~\ref{app:bad-words}; this appendix contains offending words!);
  \item removed documents that contain any non-Latin characters. This is a very strict filter; the assumption here is that we only ``trust'' a curated knowledge-database like Wikipedia to contain non-Latin script (e.g., to describe the original name of a person or place). General web data with non-Latin texts is filtered out, as a harsh quality heuristic.
\end{itemize}

While CulturaX alone is therefore filtered relatively strictly, a second step of filtering also includes the Wikipedia data. The following heuristics were calculated on the SONAR-500 corpus to serve as a baseline of ``high quality data'' and manually checked for its applicability (excluding more noisy components WRPEA, WRPED, WRUEA, WRUED, WRUEB). The following measures were taken:

\begin{itemize}
  \itemsep-0.5em
  \item removed documents where ratio of punctuation marks vs. non-whitespace characters is higher than 0.2;
  \item removed documents where ratio of uppercase vs. non-whitespace characters is higher than 0.22;
  \item removed documents where ratio of digits vs. non-whitespace characters is higher than 0.16;
  \item removed documents where the average token length is less than 2 or greater than 20.
\end{itemize}

While these filters are a useful starting point, more thorough, Dutch-specific quality filters or classifiers will be needed to ensure high-volume, high-quality datasets to efficiently train Dutch LLMs.

While only a subset of 28 billion tokens was used for Fietje, larger subsets (up to 55B tokens) of this filtered CulturaX and Wikipedia mix are also made available. The full dataset is available at \url{https://huggingface.co/datasets/BramVanroy/wikipedia\_culturax\_dutch}.

\paragraph{Training} Fietje was trained for around two weeks on four nodes of four A100 80~GB GPUs each (16 total). Note, however, that due to a CUDA configuration bug, the actual processing time should have been much shorter. Training is reproducible with the configuration file provided in Appendix~\ref{app:training-base}, also available on the aforementioned Github URL. The model weights can be found on the Hugging Face Hub.\footnote{\url{https://huggingface.co/BramVanroy/fietje-2}}

\subsection{Instruct}

Fietje itself is only capable of text completion, so to adapt it for instruction following, it was further trained with supervised finetuning on semi-structured data, namely Dutch conversations. The result of this process is Fietje Instruct.\footnote{\url{https://huggingface.co/BramVanroy/fietje-2-instruct}}

\paragraph{Data} Although large datasets for instruction following are lacking in native Dutch, a number of synthetic datasets exist. UltraChat 200K Dutch\footnote{\url{https://huggingface.co/datasets/BramVanroy/ultrachat_200k_dutch}} and No Robots Dutch\footnote{\url{https://huggingface.co/datasets/BramVanroy/no_robots_dutch}} are two datasets that were introduced by \citeasnoun{vanroy2024geitje7bultraconversational}. They were both generated with GPT-4. UltraChat 200K Dutch contains conversations that have multiple user-assistant turns. Furthermore, it was created with user accessibility in mind: the ``user'' messages in the dataset are written by GPT-4 taking on a ``persona'' such as a language learner, a critic, or a curious child. The intention is that a model finetuned on this data is better equipped to handle such diversity of users.

Belebele \cite{bandarkar-etal-2024-belebele} is a dataset intended for multiple-choice reading comprehension. For this instruction-tuning step, it was converted into the appropriate conversational format.\footnote{\url{https://huggingface.co/datasets/BramVanroy/belebele_dutch}} While small, the dataset was created by humans, ensuring its high quality.

In total, these datasets amount to 201,579 conversations.

\paragraph{Training} The instruction tuned version of Fietje was trained for around a day on four nodes of four A100 80~GB GPUs each (16 total). The training configuration can be found on Github or in Appendix~\ref{app:training-instruct}. Note that the configuration ensures that the ChatML chat template is applied.

\subsection{Chat}

As a last step in the creation process, Fietje Instruct was also ``aligned'' with preference data, specifically using the Direct Preference Optimisation algorithm \cite{rafailov2024dpo}. In this process the model is provided with a prompt and a preferred reply and an unwanted reply, and it will learn that the content, style, or way of replying of the preferred answer is something to mimic whereas the unwanted reply should be avoided.

\paragraph{Data}

Similar to the datasets for Dutch instruction tuning, no large, native Dutch preference datasets exist. Synthetic datasets, are available though. Fietje Chat\footnote{\url{https://huggingface.co/BramVanroy/fietje-2-chat}} was trained on 18,653 preference pairs taken from UltraFeedback Dutch Cleaned as well as Orca DPO Pairs Dutch Cleaned, both described in \citeasnoun{vanroy2024geitje7bultraconversational}. The \texttt{dpo\_hq} subset was taken from UltraFeedback\footnote{\url{https://huggingface.co/datasets/BramVanroy/ultra_feedback_dutch_cleaned}}, which contains Dutch prompts that were answered by GEITje 7B Ultra and GPT-4. GPT-4 was then asked to rate the responses on the quality of the Dutch language, their helpfulness and conciseness -- having LLMs rate their own responses is an intriguing but common practice. The best rated response, GEITje or GPT-4 was then used as the ``preferred'' response and the other as the unwanted one. Similarly, for Orca DPO Pairs\footnote{\url{https://huggingface.co/datasets/BramVanroy/orca_dpo_pairs_dutch_cleaned}}, the \texttt{dpo\_all} subset was used for training. This subset was not rated, so GPT-4 was always selected as the preferred response and GEITje as the worst response.

\paragraph{Training} Since the DPO training set was quite small, only one A100 80GB was used. One training run took around 9 hours. However, similar to the experience of \citeasnoun{vanroy2024geitje7bultraconversational}, it was found that DPO's $beta$ parameter was difficult to tune while avoiding hallucinations or catastrophic forgetting. This parameter controls how close the model should stick to its starting point, or how far it should move away. The final beta value of $0.2$ was selected after hyperparameter tuning. The full configuration is given in Appendix~\ref{app:training-chat} and on Github.

\section{Evaluation}

In this section we will evaluate Fietje and its instruction and chat variants on a number of benchmarks, and compare its results with other language models, some of which are also specifically tailored to Dutch, others which are multilingual. Plenty of models are given to highlight a few insights when it comes to model release date (multilingual performance has improved significantly since Fietje's release), model sizes, and the impact of adaptation to Dutch.

Crucially, it is important to emphasize that Dutch, and many non-English languages, has been struggling with LLM evaluation. As discussed before in \cite{vanroy2023languageresourcesdutchlarge,vanroy2024geitje7bultraconversational}, most existing benchmarks are translated, are ``simply'' classification problems that do not measure the fluency of the model, and/or are not localized to the intricacies of Dutch language culture. This is discussed in more detail in Section~\ref{sec:benchmark-overview}.

\subsection{Framework}

While evaluation frameworks for generative large language models exist, such as ScandEval \cite{nielsen2023scandeval} and the LM Evaluation Harness \cite{eval-harness}, they either did not provide the benchmarks (for Dutch) that I wanted to emphasize, did not allow ease-of-customizability with configuration files, or did not include functionality such as speed and fertility tests that I needed. Notably, while ScandEval specifically is a useful resource for comparing models, it only evaluates on a limited number of tasks, and -- crucially -- some tasks are only evaluated on parts of the test set rather than the whole test set to save on computational costs, which is not always clear as a viewer of the ``leaderboard''. Therefore, an extensible evaluation framework is made available accompanying this paper to reproduce all results presented here. The benchmark code (including fertility and throughput calculation) and accompanying benchmark configuration files are available at \url{https://github.com/BramVanroy/clin34-benchmarks} for full reproducibility.

To enable efficient guided generation in the benchmarks, Outlines was used as a backbone of the benchmarking suite \cite{willard2023outlines}. This library enables you to constrain a model's output by regular expressions, JSON schema, or simply a list of options. In zero-shot benchmarks like the ones discussed here, that means that the model will always predict one of the allowed labels and is not able to hallucinate labels or responses. For instance, for a sentiment analysis task the prompt may be in the line of ``Would you say that the following book review is `positive' or `negative'?'' and in the benchmarking code we can explicitly state that the model is only allowed to return `positive' or `negative', leading to clean benchmarking results that do not require any post-processing.

In the benchmarks, sampling \textit{was} enabled (no top p or top k selection, and temperature 1). That means that the model's predictions are limited thanks to Outlines to the list of valid labels (e.g. `positive' and `negative') but that due to sampling, the predicted outcome are still randomly sampled, but weighted according to their initial probability. Since each prediction is run multiple times, the confidence interval will therefore be an important metric to consider. If the model is 99\% confident that `positive' is the right answer, then the confidence interval will be small because even with random sampling, `positive' will be the most picked answer. However, if the model is only 50\% confident, then the confidence interval would likely be much larger since the random sampling will lead to a 50/50 choice on the labels. Therefore, large confidence intervals will be indicative of a model's poor confidence, as the name implies. Each benchmark was run five times to compute the confidence intervals.

Benchmarks were run in bfloat16 precision with flash attention 2 enabled \cite{dao2024flashattention}. All benchmarks were run on four RTX 3090 with 24~GB of VRAM each.

\subsection{Model overview}

To compare the Fietje models, a number of different large language models were selected for comparison. Many of these were released after Fietje -- so the expectation is that newer models will perform better -- but including them highlights the rapid changes in the (multilingual) LLM world. In Table~\ref{tab:model-overview}, notable model characteristics are given. The columns are described in the table's caption. In the Wikipedia columns, fertility (the average ratio of how many subword tokens are needed to encode a word) is calculated on full Dutch Wikipedia (cleaned dump of November 2023)\footnote{\url{https://huggingface.co/datasets/wikimedia/wikipedia/viewer/20231101.nl}}, whereas the tokens-per-second (wiki $tps$) and processing time in seconds (wiki $s$) were calculated on the first 10,000 documents of Dutch Wikipedia. The data and training transparency columns are to indicate how reproducible a model is: is the data specifically described and publicly available, and is the training code or configuration provided?

\paragraph{Phi 2 and derivatives} Phi 2 \cite{javaheripi2023phi2} serves as the base model for Fietje. It was discussed before in Section~\ref{sec:creation}. Phi 2 and its derivatives are the smallest model in the list in terms of parameter count. Its tokenizer is a bit worse than other models in terms of fertility: on average 2.05 subword tokens are needed to encode a single Dutch word, highlighting that it was intended for English. Still, given its small size it is the fastest model in the list: Phi 2 related models can process most tokens per second and are also the fastest in processing time of Wikipedia. While all Fietje models are fully reproducible (public code, public data), that is not the case for Phi 2, which is not specific about the data or training regimen used.

\paragraph{Mistral v0.1} Mistral v0.1 \cite{jiang2023mistral7b} is a 7.2-billion-parameter language model known for its performance in monolingual English benchmarks at the time. Mistral v0.1 was not trained on multilingual data and lacks optimizations for the Dutch language according to their model card, which only mentions English. Consequently, its fertility on Dutch text is relatively high, averaging 1.97 subword tokens per word. Combined with its large parameter count, this leads to the slowest throughput in terms of Wikipedia tokens per second ($tps$) among the models compared. Despite its technical advancements at the time, Mistral v0.1 suffers from a lack of transparency regarding its training data and procedures. While the model architecture and code are available under an open-source license, the exact data sources and training configurations remain undisclosed, limiting the reproducibility of its results.

Prior efforts were taken to make Mistral v0.1 more fluent in Dutch by continue-pretraining the model without changing the tokenizer. The first successful attempt in this regard was the creation of GEITje \cite{rijgersberg2023geitje}, a continued-pretrained version of Mistral on 10 billion Dutch tokens which have been clearly described though not all publicly available. The training code is publicly available. GEITje 7B was later extended by \cite{vanroy2024geitje7bultraconversational} into ``GEITje 7B Ultra'' by finetuning the model on new, open instruction datasets as well as aligning the model with preference dating using Direct Preference Optimisation \cite{rafailov2024dpo}. More recently, another Mistral v0.1 continued-pretrained model was launched called Boreas 7B\footnote{\url{https://huggingface.co/yhavinga/Boreas-7B-chat}} and its instruct version called Boreas 7B Chat (which is an instruction-tuned model, not a preference-tuned model). It was trained on a mix of 10 billion English and Dutch tokens with an extensive finetuning phase for instruction-tuning with another 4.7B tokens, much more than the instruction tuning for GEITje. The datasets are a combination of public and private datasets that include lots of Dutch literature, news and educational materials. Training code is not available but training hyperparameters are discussed in the model card.

\paragraph{Tweety} Although the Dutch 7B Tweety model \cite{remy2024transtokenization} is also a derivative of Mistral v0.1, it is special. As described in Section~\ref{sec:related-work}, Tweety has an updated tokenizer and vocabulary to tailor it better to the Dutch language, as is clear from its outstanding low fertility. This is clear in the architecture as well as in its efficiency: while it is a bit larger than other Mistral derivatives (7.4B vs 7.2B parameters), leading to a slower throughput (wiki $tps$), it is still much faster in processing Wikipedia (wiki $s$). In other words: because its tokenizer is optimized for Dutch, it needs fewer subword tokens to encode the same information, and therefore can process the same text much more efficiently. After updating the tokenizer, the model was ``finetuned'' on 400 million Dutch tokens from the mC4 corpus \cite{Raffel2019ExploringTL} to adapt to the new embeddings, although the exact subset is not known. The code for updating the tokenizer and model is available\footnote{\url{https://github.com/LAGoM-NLP/transtokenizer}} but the finetuning code is not.

\paragraph{Phi 3.5} The Phi 3 and Phi 3.5 models \cite{abdin2024phi3} represent a significant evolution in the Phi family, with Phi 3.5 incorporating multilingual data during mid-training. However, the exact composition of languages in the training corpus is not disclosed -- in fact, no languages are explicitly listed, resulting in a lack of transparency regarding the data sources and training procedures. Despite this, these models exhibit improved tokenizer performance compared to Phi 2, with a fertility of 1.89 subword tokens per Dutch word, indicating a better adaptation to the Dutch language. This enhancement contributes to more efficient processing of Dutch text. In this paper we discuss Phi 3.5 Mini Instruct, a 3.3B version of Phi 3.5.

\paragraph{Qwen 2.5} Qwen 2.5 \cite{qwen2.5} is a series of multilingual language models trained on over 29 languages. In this paper, for comparison with Fietje, the 3.1-billion-parameter instruction variant is selected. Although Dutch is not explicitly mentioned among the language list (in fact the 29 languages are not listed in full), the model demonstrates a commendable tokenizer fertility of 1.82 subword tokens per Dutch word, surpassing both Phi 2 and Phi 3.5. This efficiency, coupled with its relatively small size, enables Qwen 2.5 to process Dutch Wikipedia text with fast speed. However, the model's transparency is limited, as details regarding its training data and code are not publicly available, hindering reproducibility and understanding of the model creation.

\paragraph{Llama 3.2} The Llama 3.2 3B Instruct model is part of Meta's Llama 3 series \cite{grattafiori2024llama3}, which supports multilingual training across eight languages. Although Dutch is not explicitly listed, the tokenizer exhibits strong performance on Dutch text, second only to Tweety's Dutch-optimized tokenizer. With a fertility notably low for a general-purpose model, Llama 3.2 processes Dutch Wikipedia almost as efficiently as Phi 2, despite being 14\% larger in parameter count. This balance of size and efficiency results in high throughput for Dutch text processing. The Llama 3 paper stands out for its transparency, offering extensive insights into model creation, data curation, and technical training details. While the exact data mix and training code are not provided, the paper describes a rich blend of code, math, multilingual, and filtered web data, including references to specific public datasets.

\begin{landscape}
\vfill 
\begin{table}[]
\centering
\resizebox{\linewidth}{!}{
\begin{tabular}{l|l|l|l|l|l|l|l|l|l|l}
\thead{name} & \thead{date} & \thead{size} & \thead{type} & \thead{Dutch-specific} & \thead{data\\transparency} & \thead{training\\transparency} & \thead{finetuned from} & \thead{wiki\\fertility} & \thead{wiki $tps$} & \thead{wiki $s$} \\ \hline 
\textbf{fietje-2b} & 4/24 & 2.8B & base & yes & yes & yes & phi-2 & 2.05 & 9501.41 ± 0.66 & 440.41 ± 0.03 \\
\textbf{fietje-2b-chat} & 4/24 & 2.8B & chat & yes & yes & yes & fietje-2b-instruct & 2.05 & 9501.41 ± 0.66 & 440.41 ± 0.03 \\
\textbf{fietje-2b-instruct} & 4/24 & 2.8B & instruct & yes & yes & yes & fietje-2b & 2.05 & 9501.70 ± 4.72 & 440.40 ± 0.22 \\
\textbf{GEITje-7B-ultra} & 1/24 & 7.2B & chat & yes & yes & yes & GEITje-7B & 1.97 & 4035.27 ± 0.64 & 999.42 ± 0.16 \\
\textbf{Llama-3.2-3B-Instruct} & 9/24 & 3.2B & instruct & no & partly & partly & Llama-3.2-3B & 1.74 & 7884.63 ± 0.36 & 451.97 ± 0.02 \\
\textbf{phi-2} & 12/23 & 2.8B & base & no & no & no & none & 2.05 & 9631.95 ± 16.12 & 434.44 ± 0.73 \\
\textbf{Phi-3.5-mini-instruct} & 8/24 & 3.8B & instruct & underspecified & no & no & none & 1.89 & 6633.85 ± 0.68 & 584.14 ± 0.06 \\
\textbf{Mistral-7B-Instruct-v0.1} & 9/23 & 7.2B & instruct & no & unclear\footnote{Their paper suggests that they only trained on public data but they do not specify which sets} & no & none & 1.97 & 4027.81 ± 1.14 & 1001.27 ± 0.28 \\
\textbf{Mistral-7B-v0.1} & 9/23 & 7.2B & base & no & no & no & Mistral-7B-v0.1 & 1.97 & 4046.46 ± 0.67 & 996.66 ± 0.16 \\
\textbf{Qwen2.5-3B-Instruct} & 9/24 & 3.1B & instruct & underspecified & no & no & Qwen2.5-3B & 1.82 & 8094.26 ± 0.53 & 459.94 ± 0.03 \\
\textbf{GEITje-7B} & 12/23 & 7.2B & base & yes & partly & yes & Mistral-7B-v0.1 & 1.97 & 4021.61 ± 1.64 & 1002.82 ± 0.41 \\
\textbf{tweety-7b-dutch-v24a} & 5/24 & 7.4B & base & yes & yes & partly & Mistral-7B-v0.1 & 1.41 & 3979.88 ± 2.12 & 728.56 ± 0.39 \\
\textbf{Boreas-7B} & 4/24 & 7.2B & base & yes & partly & partly & Mistral-7B-v0.1 & 1.97 & 4032.05 ± 15.28 & 1000.23 ± 3.78 \\
\textbf{Boreas-7B-chat} & 4/24 & 7.2B & instruct & yes & partly & partly & Boreas-7B & 1.97 & 4034.36 ± 0.69 & 999.65 ± 0.17
\end{tabular}
}
\caption{Overview of benchmarked models. \textbf{Dutch-specific}: did the model undergo (re-)training specifically for Dutch? \textbf{data/training transparency}: are the data and training procedure described in detail (reproducible) and is the data and training code publicly available? \textbf{wiki fertility}: how many tokens are needed on average to encode one word, calculated on full Dutch Wikipedia. Lower = more efficient. \textbf{wiki tps}: Tokens-per-second throughput on first 10,000 Wikipedia documents. How many tokens can the model process per second. \textbf{wiki s}: Processing time of first 10,000 Wikipedia documents. Lower = faster. \textbf{wiki tps} and \textbf{wiki s} were calculated on an isolated RTX 3090 in bfloat16 with Flash Attention 2 enabled. Batch size was 1 and all models' maximum context length was used, or 8192 at most. The reported mean metrics and their CI are based on the results of three runs.}
\label{tab:model-overview}
\end{table}
\end{landscape}

\subsection{Benchmark overview}\label{sec:benchmark-overview}

The following benchmarks are considered to cover different aspects of LLM capabilities: ARC (reasoning), DBRD (sentiment analysis), Dutch CoLA (grammar/linguistic acceptability), Global MMLU (language understanding and world knowledge), XLWIC-NL (word sense disambiguation). Argumentation for this selection of benchmarks, and why others were not included, is given in Section~\ref{sec:limitation-eval}.

\paragraph{ARC} The AI2 ARC (AI2 Reasoning Challenge) dataset \cite{clark2018arc} consists of 7,787 natural, grade-school science questions in advanced reasoning beyond simple fact retrieval. The dataset is divided into two parts: an Easy Set (5,197 questions) and a Challenge Set (2,590 questions). This English challenge dataset was translated with GPT-3.5-turbo to a multitude of languages, including Dutch, by \citeasnoun{lai-etal-2023-okapi}. The dataset is presented to a model as multiple choice questions.

\paragraph{DBRD} The Dutch Book Reviews Dataset \cite{dbrd} is a sentiment analysis dataset based on actual book reviews taken from the website hebban.nl. Each review was accompanied by a score out of five. The reviews with a score of 4 and 5 were labeled as `positive' while those marked 1 or 2 were labeled as `negative'. The dataset contains 2,224 test samples.

\paragraph{Dutch CoLA} The Dutch CoLA dataset\footnote{\url{https://huggingface.co/datasets/GroNLP/dutch-cola}} \cite{gronlp2024cola} is a linguistic acceptability corpus for Dutch, following the structure of the English CoLA dataset \cite{warstadt2019cola}. It comprises sentences labeled as grammatically correct or incorrect, derived from expert-annotated examples found in published Dutch grammar literature. The dataset contains 2,400 test sentences.

\paragraph{Global MMLU} The Global MMLU (Massive Multitask Language Understanding) benchmark \cite{singh2024globalmmlu} is an extension and improvement of the original English MMLU \cite{hendrycks2021mmlu}, designed to evaluate LLMs across multiple languages, including Dutch. This benchmark measures a model's ability to perform on a wide range of academic and professional knowledge tasks spanning STEM, humanities, social sciences, and more, with questions ranging in difficulty from elementary to expert levels. Questions are formulated as multiple choice questions.

While the original MMLU focuses solely on English, the Global MMLU provides translated and culturally verified versions of the dataset. The Dutch portion was generated by first translating the English MMLU questions using machine translation and subsequently improving the quality through human post-editing. Note that this improved MMLU version is different from earlier machine-translated versions of MMLU by \citeasnoun{lai-etal-2023-okapi}.

\paragraph{XLWIC-NL} The multilingual Word-in-Context dataset \cite{raganato-etal-2020-xlwic} is intended for word sense disambiguation of nouns and verbs. The Dutch portion is derived from the Dutch WordNet, which provides curated sense inventories and example usages, ensuring reliability in distinguishing different word senses. A model is presented with a target word and two sentences where the word (or its conjunction) is used. The model should then predict whether the meaning of the word is identical or different in the two sentences. The test set includes 1,004 test instances.

\subsection{Results}

In this section, the performance of the models on the selected benchmarks is presented and discussed from different points of view. While the focus of this paper lies on Fietje and its derivatives, the benchmark results also give food for thought about other models and the tasks themselves.

\begin{landscape}
\vfill 
\begin{table}[]
\centering
\resizebox{\linewidth}{!}{
\begin{tabular}{l|rr|rr|rr|rr|rr||r}
 & \thead{Global\\MMLU} & \thead{Global\\MMLU\\rank} & \thead{DBRD} & \thead{DBRD\\rank} & \thead{Dutch\\CoLA} & \thead{Dutch\\CoLA\\rank} & \thead{ARC} & \thead{ARC\\rank} & \thead{XLWIC} & \thead{XLWIC\\rank} & \thead{\textbf{median rank}} \\ \hline \hline
\textbf{Phi-3.5-mini-instruct} & 48.34 ± 0.10 & 2 & 92.31 ± 0.13 & 2 & 58.43 ± 0.09 & 2 & 65.31 ± 0.22 & 2 & 37.39 ± 0.31 & 8 & 1.0 \\ \hline
\textbf{Qwen2.5-3B-Instruct} & \textbf{50.33 ± 0.14} & 1 & 91.70 ± 0.15 & 3 & \textbf{63.74 ± 0.17} & 1 & \textbf{66.97 ± 0.45} & 1 & 36.05 ± 0.38 & 12 & 2.0 \\ \hline
\textbf{Boreas-7B-chat} & 44.93 ± 0.15 & 3 & \textbf{94.38 ± 0.27} & 1 & 52.87 ± 0.42 & 4 & 59.88 ± 0.66 & 3 & 33.78 ± 0.34 & 14 & 3.5 \\ \hline
\textbf{Llama-3.2-3B-Instruct} & 35.59 ± 0.22 & 4 & 59.74 ± 0.97 & 8 & 55.35 ± 1.45 & 3 & 42.80 ± 0.93 & 4 & 42.72 ± 0.76 & 3 & 3.5 \\ \hline
\textbf{Mistral-7B-Instruct-v0.1} & 32.30 ± 0.38 & 5 & 79.73 ± 0.67 & 5 & 40.29 ± 0.75 & 14 & 36.72 ± 1.06 & 5 & 40.03 ± 0.53 & 4 & 5.0 \\ \hline
\textbf{GEITje-7B-ultra} & 24.39 ± 0.18 & 12 & 90.00 ± 0.37 & 4 & 46.57 ± 0.59 & 9 & 29.10 ± 0.56 & 8 & \textbf{44.45 ± 0.79} & 1 & 6.0 \\ \hline
\textbf{tweety-7b-dutch-v24a} & 27.36 ± 0.32 & 7 & 40.22 ± 1.04 & 14 & 51.27 ± 1.00 & 5 & 29.46 ± 1.25 & 7 & 43.23 ± 1.07 & 2 & 7.0 \\ \hline
\textbf{fietje-2b-chat} & 26.36 ± 0.25 & 9 & 58.78 ± 0.58 & 9 & 45.45 ± 0.64 & 10 & 31.56 ± 0.78 & 6 & 39.24 ± 0.94 & 5 & 8.0 \\ \hline
\textbf{Boreas-7B} & 27.02 ± 0.63 & 8 & 70.33 ± 1.12 & 6 & 49.34 ± 0.51 & 6 & 26.19 ± 0.85 & 12 & 37.24 ± 0.98 & 10 & 9.5 \\ \hline
\textbf{fietje-2b-instruct} & 24.93 ± 0.27 & 11 & 51.38 ± 0.76 & 12 & 49.31 ± 0.78 & 7 & 28.70 ± 0.82 & 9 & 38.61 ± 1.00 & 6 & 9.5 \\ \hline
\textbf{Mistral-7B-v0.1} & 27.51 ± 0.15 & 6 & 63.69 ± 0.85 & 7 & 48.00 ± 0.51 & 8 & 26.82 ± 0.85 & 11 & 37.27 ± 0.88 & 9 & 11.0 \\ \hline
\textbf{GEITje-7B} & 25.12 ± 0.37 & 10 & 46.28 ± 0.78 & 13 & 43.67 ± 0.72 & 11 & 27.61 ± 1.30 & 10 & 37.64 ± 0.75 & 7 & 12.0 \\ \hline
\textbf{phi-2} & 20.82 ± 0.28 & 14 & 51.45 ± 0.66 & 11 & 42.29 ± 0.75 & 12 & 18.07 ± 0.52 & 14 & 36.55 ± 0.97 & 11 & 13.0 \\ \hline
\textbf{fietje-2b} & 24.09 ± 0.43 & 13 & 52.44 ± 1.23 & 10 & 41.41 ± 0.37 & 13 & 24.44 ± 0.89 & 13 & 34.28 ± 0.70 & 13 & 14.0
\end{tabular}
}
\caption{Benchmark results, showing weighted F1 score and the 95\% confidence interval (obtained by running each benchmark five times on each model). Models' ranks are also given, although they should be taken with a grain of salt considering overlapping confidence intervals. The last column illustrates the final median ranking across all benchmarks.}
\label{tab:benchmark-results}
\end{table}
\end{landscape}

\begin{figure}[ht]
    \centering
    \begin{subfigure}[b]{0.45\textwidth}
        \centering
        \includegraphics[width=\textwidth]{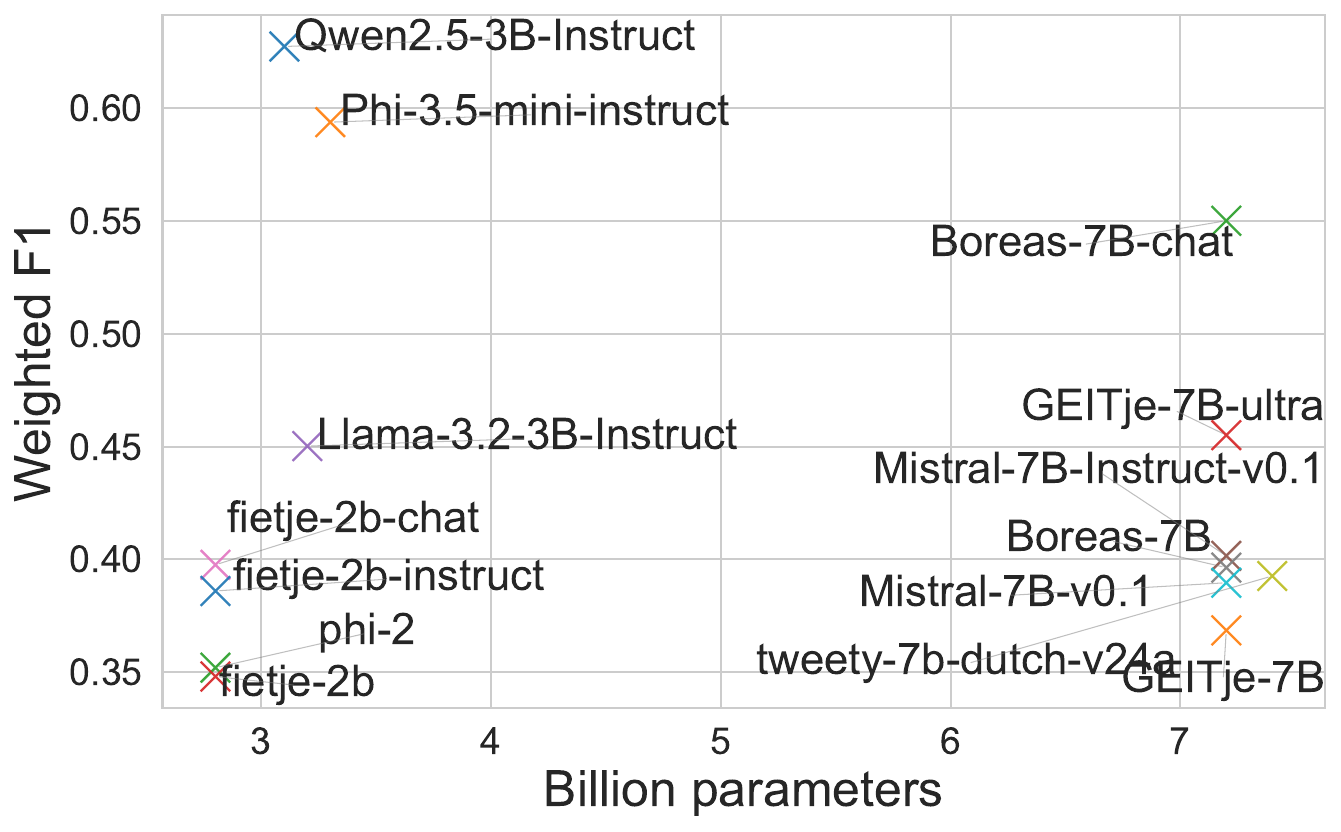}
        \caption{All models}
        \label{fig:size-all}
    \end{subfigure}
    \hfill
    \begin{subfigure}[b]{0.45\textwidth}
        \centering
        \includegraphics[width=\textwidth]{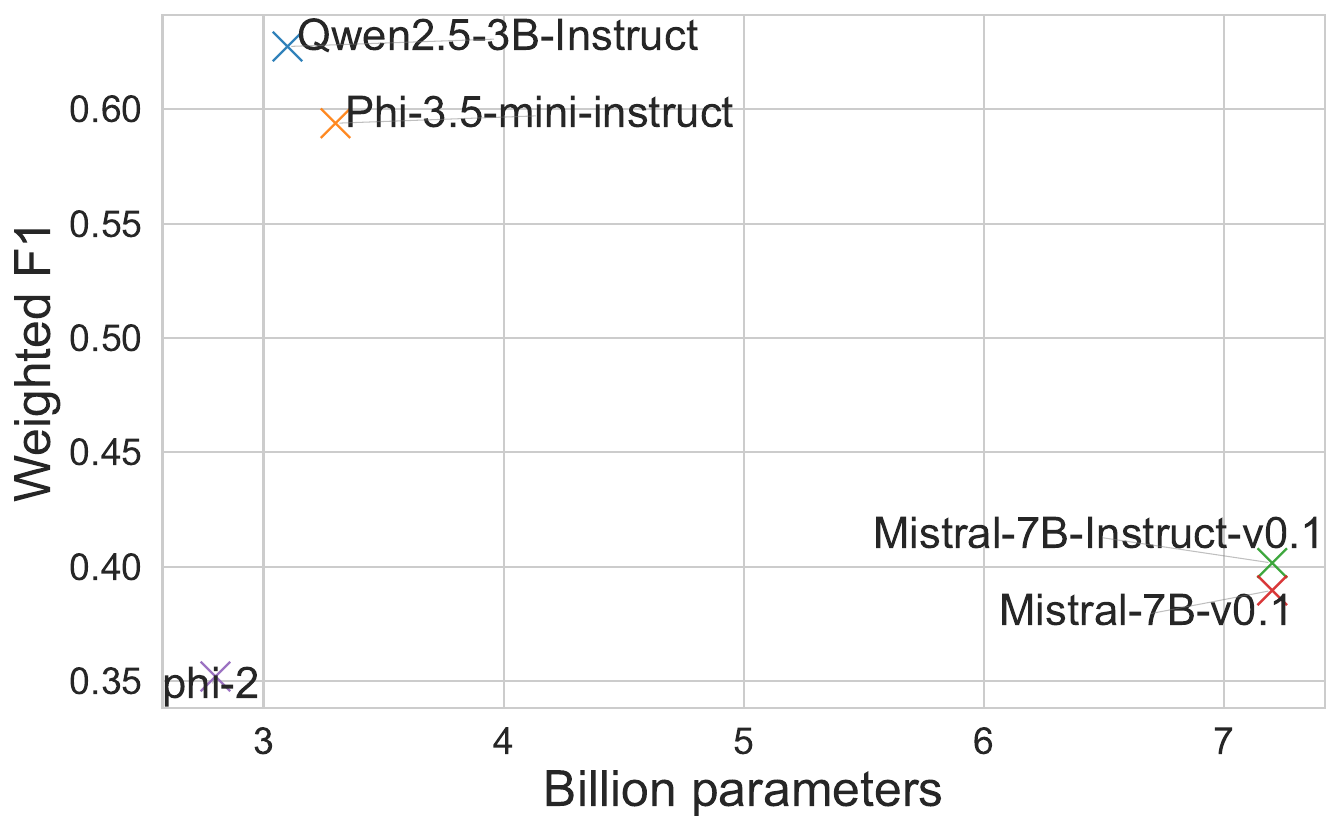}
        \caption{Without modified models}
        \label{fig:size-unmodified}
    \end{subfigure}
    \caption{Model size vs. median performance}
    \label{fig:size}
\end{figure}

Table~\ref{tab:benchmark-results} can be analysed in a number of ways:

\paragraph{Impact of model size and release date} While a realistic expectation would be that larger models perform better overall, this is not the case, as visualized in Figure~\ref{fig:size-all}. Even limiting the model selection to only those models that were not specifically adapted to Dutch (Fig.~\ref{fig:size-unmodified}), no trends are visible: larger models like Mistral 7B are outperformed by much smaller ones like Qwen 2.5 3B. Detailed visualizations for the relationship between model size and individual task performance are given in Appendix\ref{app:size-v-perf}.

\begin{figure}[ht]
    \centering
    \begin{subfigure}[b]{0.45\textwidth}
        \centering
        \includegraphics[width=\textwidth]{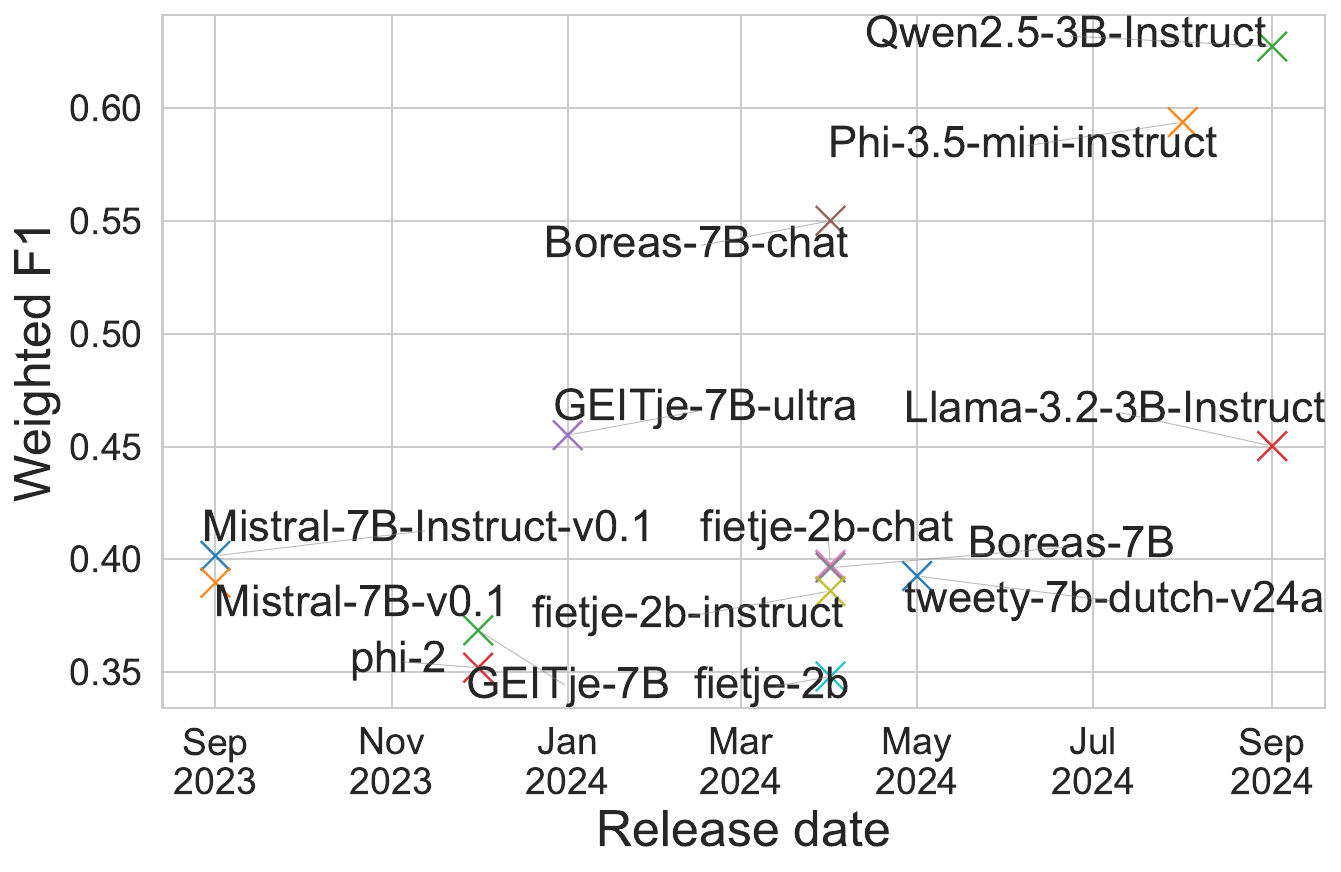}
        \caption{All models}
        \label{fig:date-all}
    \end{subfigure}
    \hfill
    \begin{subfigure}[b]{0.45\textwidth}
        \centering
        \includegraphics[width=\textwidth]{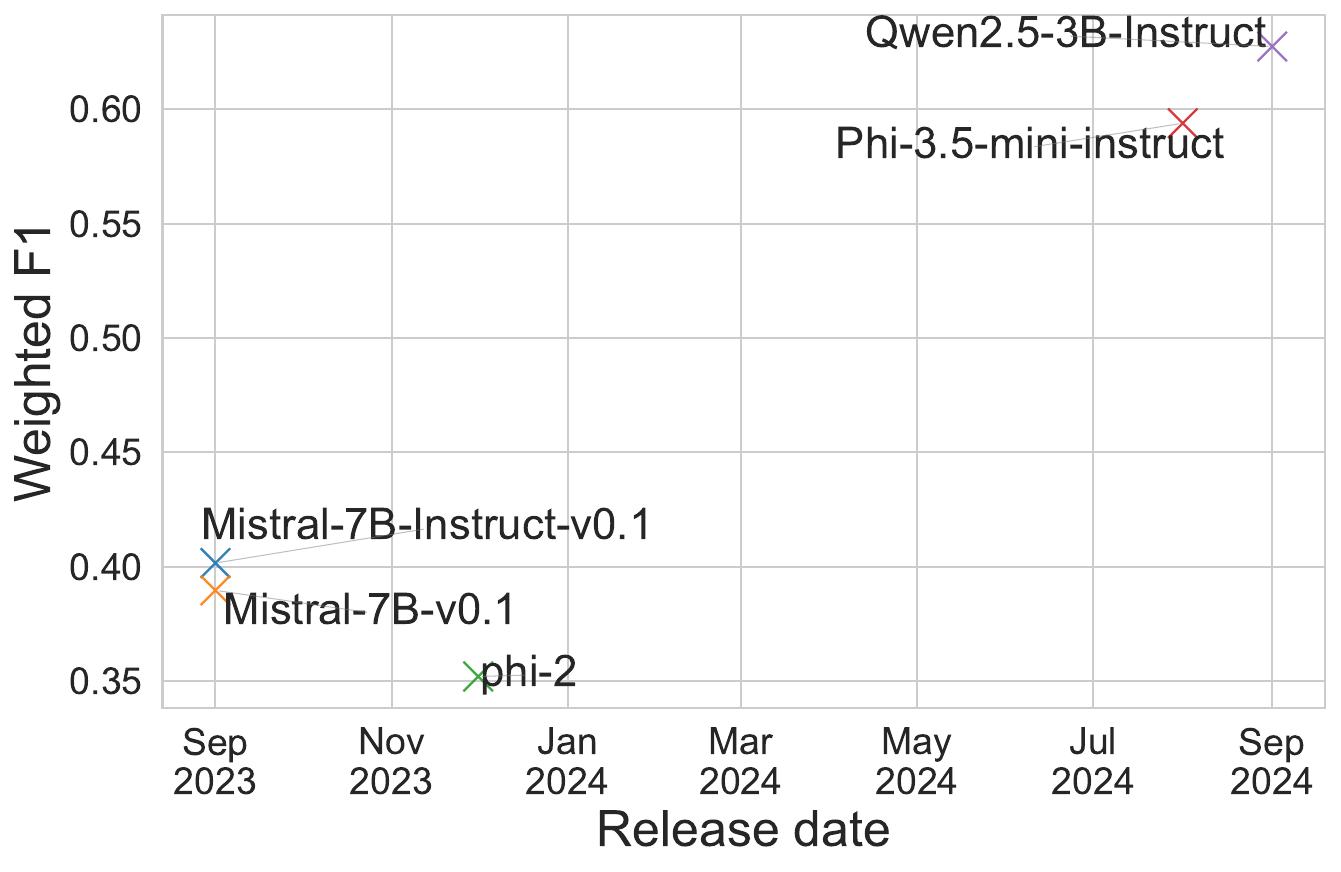}
        \caption{Without modified models}
        \label{fig:date-unmodified}
    \end{subfigure}
    \caption{Model release date vs. median performance}
    \label{fig:date}
\end{figure}

However, when considering the release date of the models (Fig.~\ref{fig:date-all}), it becomes clear that more recent models tend to have an advantage over older models (considering their respective model sizes). This becomes very clear when focusing on the un-adapted models in Figure~\ref{fig:date-unmodified}, where newer models like Phi 3.5 and Qwen 2.5 have an edge over Phi 2, and even over the much larger but older Mistral v0.1 models. Appendix~\ref{app:date-v-perf} plots the model release date on their performance of specific tasks for the interested reader.

\paragraph{Note on confidence intervals and rank} First and foremost it must be acknowledged that many models exhibit tight, sometimes even overlapping, confidence intervals, suggesting that performance differences in some tasks are marginal between certain models. While ranking models on their performance provides a quick way of putting them on a leaderboard, the ranks should be interpreted with caution. For instance, in the DBRD (sentiment analysis) benchmark, Phi 3.5 and Qwen 2.5 have confidence intervals that slightly overlap, making their ranking potentially interchangeable.

\begin{figure}[ht]
    \centering
    
    \begin{subfigure}[b]{0.45\textwidth}
        \centering
        \includegraphics[width=\textwidth]{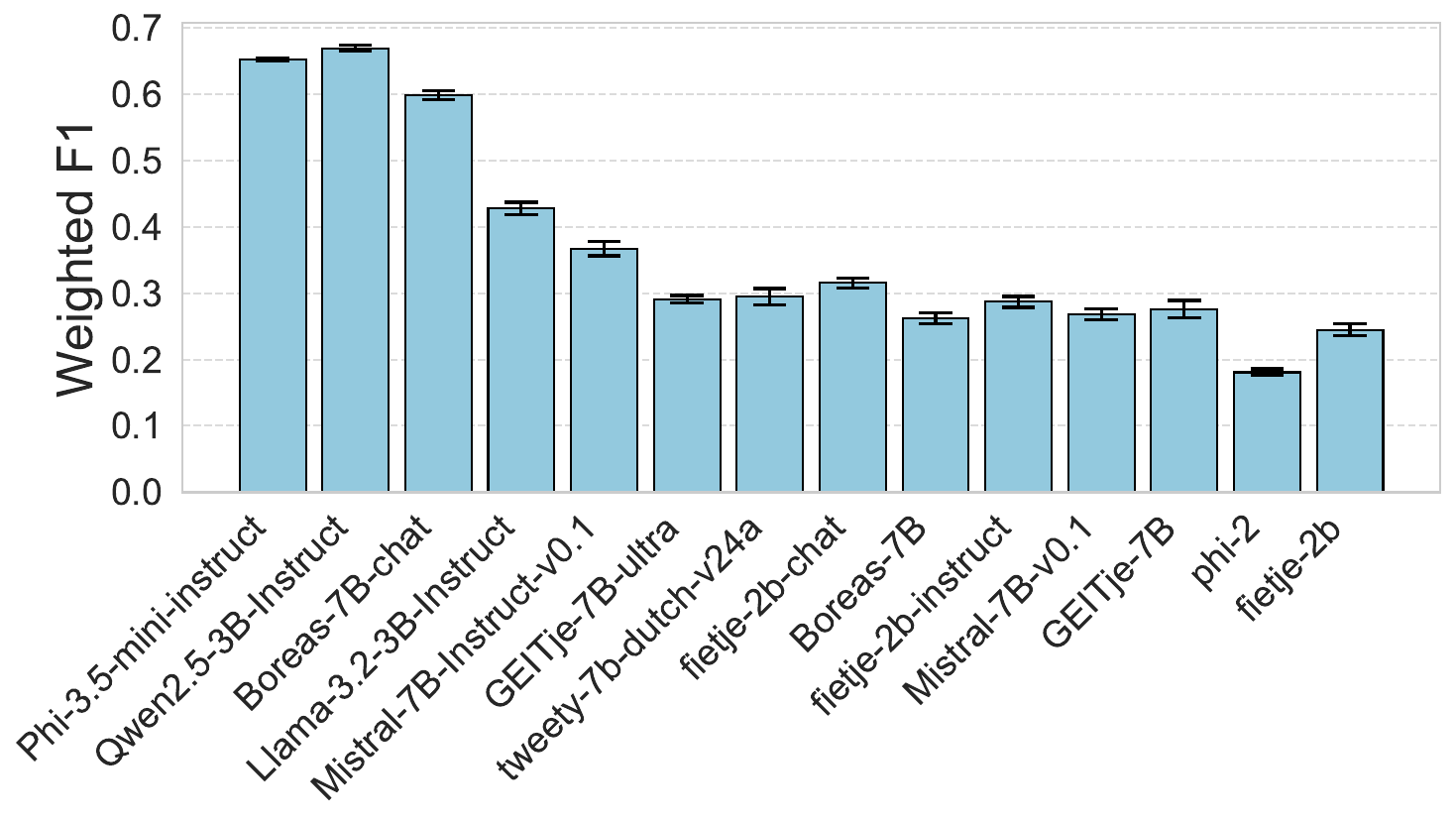}
        \caption{ARC}
        \label{fig:barplots-arc}
    \end{subfigure}
    \hfill
    \begin{subfigure}[b]{0.45\textwidth}
        \centering
        \includegraphics[width=\textwidth]{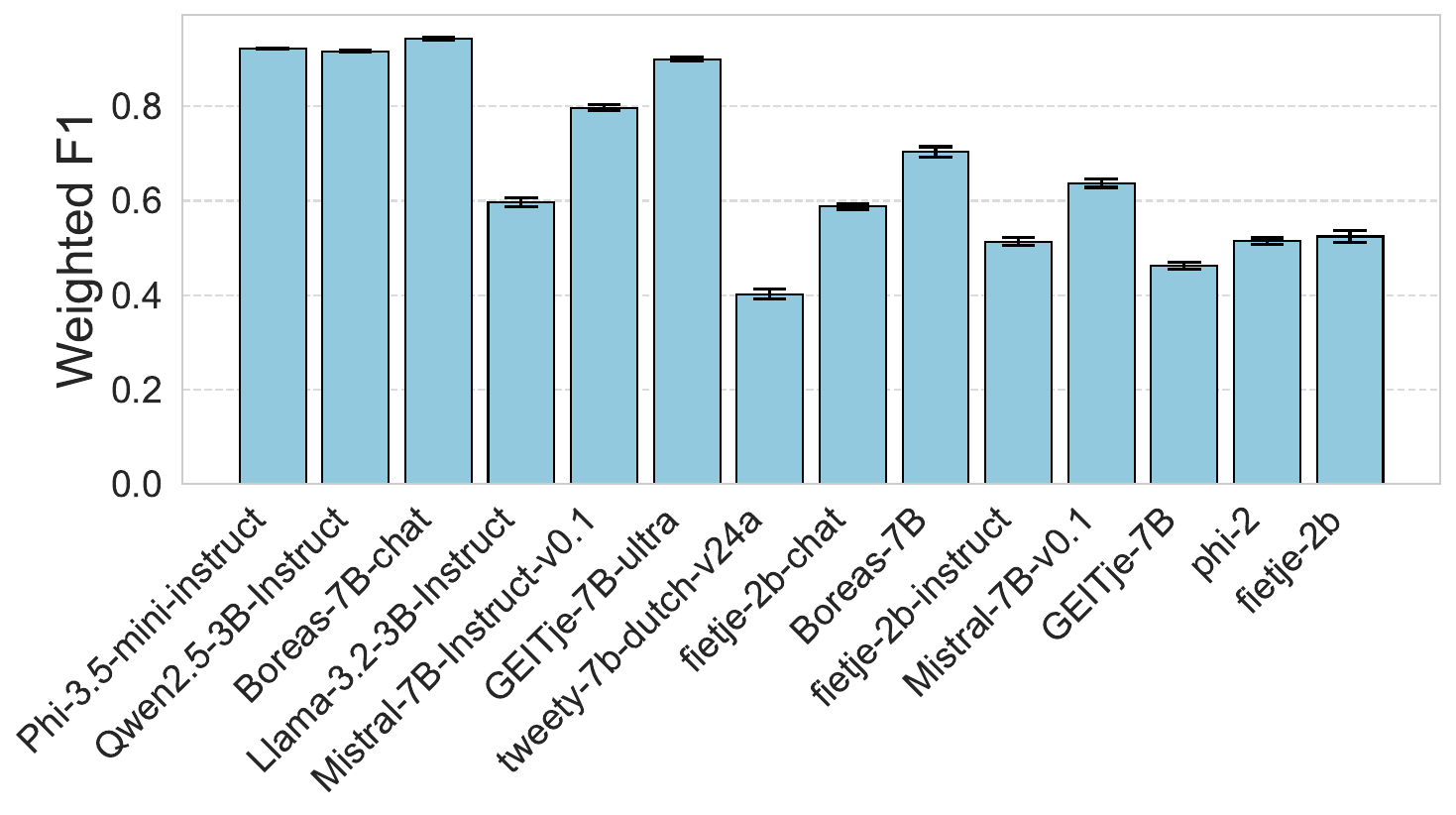}
        \caption{DBRD}
        \label{fig:barplots-dbrd}
    \end{subfigure}
    
    \vspace{0.5cm}

    \begin{subfigure}[b]{0.45\textwidth}
        \centering
        \includegraphics[width=\textwidth]{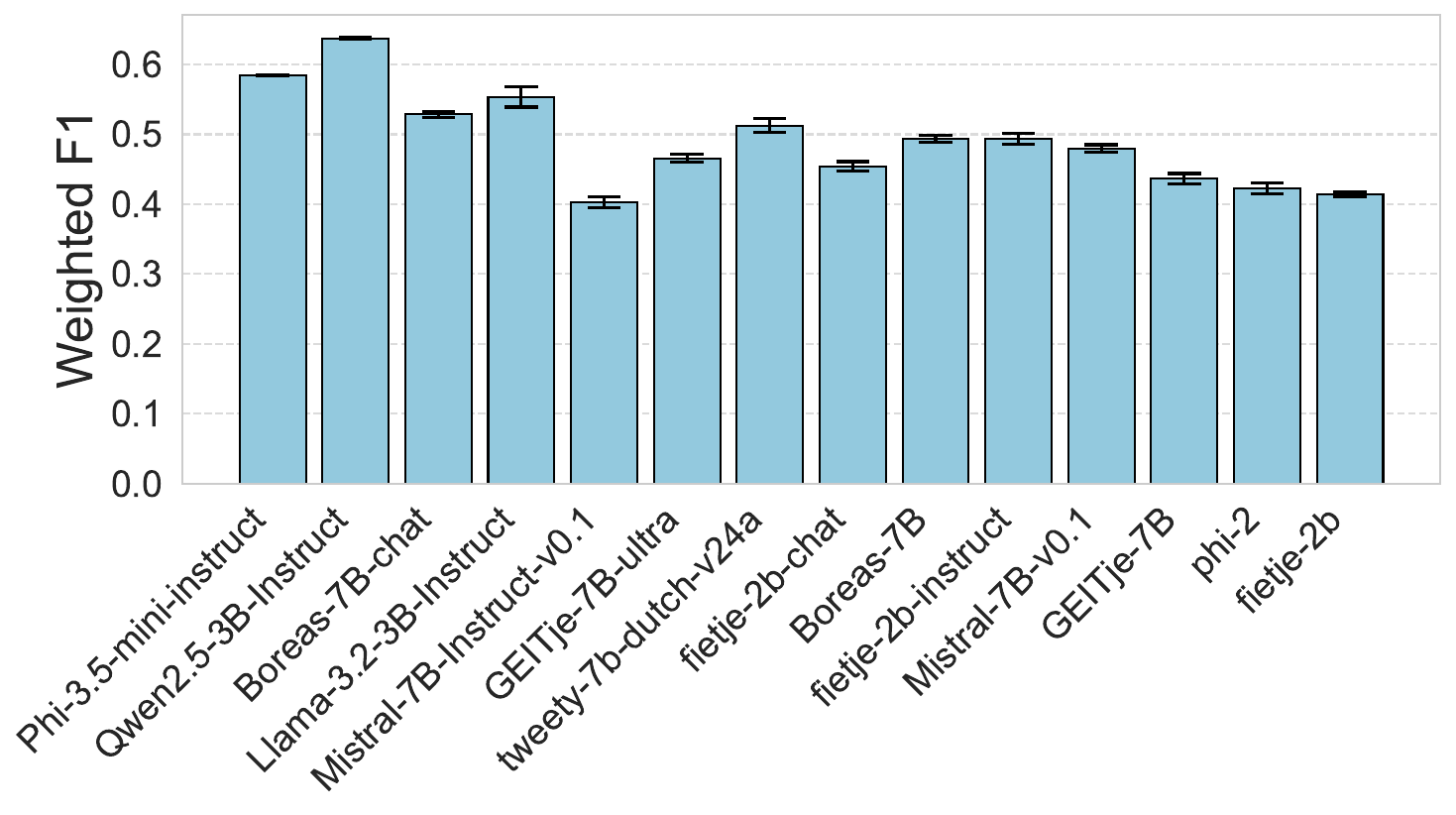}
        \caption{Dutch CoLA}
        \label{fig:barplots-dutch_cola}
    \end{subfigure}
    \hfill
    \begin{subfigure}[b]{0.45\textwidth}
        \centering
        \includegraphics[width=\textwidth]{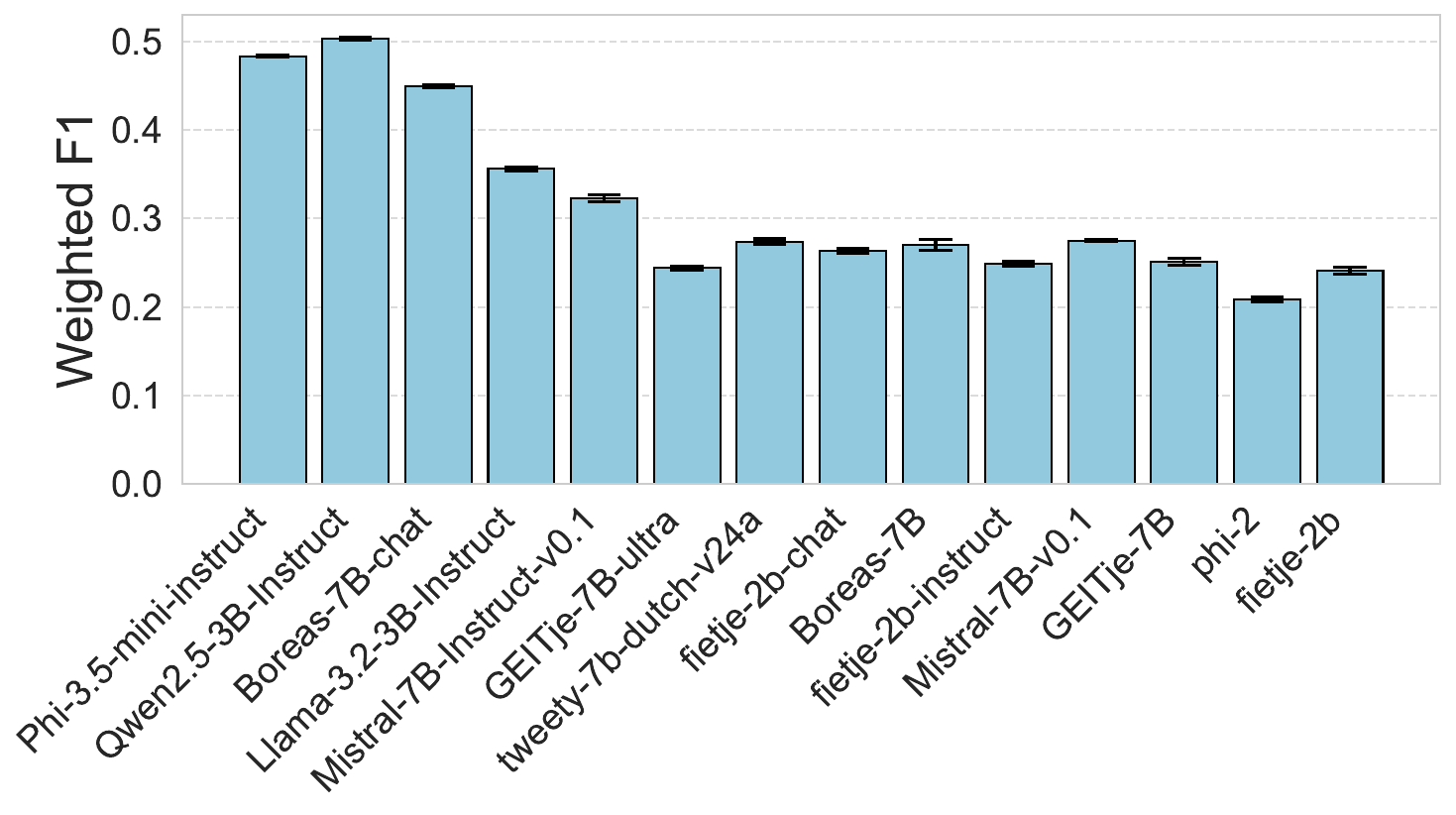}
        \caption{Global-MMLU}
        \label{fig:barplots-global_mmlu}
    \end{subfigure}

    \vspace{0.5cm}

    \begin{subfigure}[b]{0.45\textwidth}
        \centering
        \includegraphics[width=\textwidth]{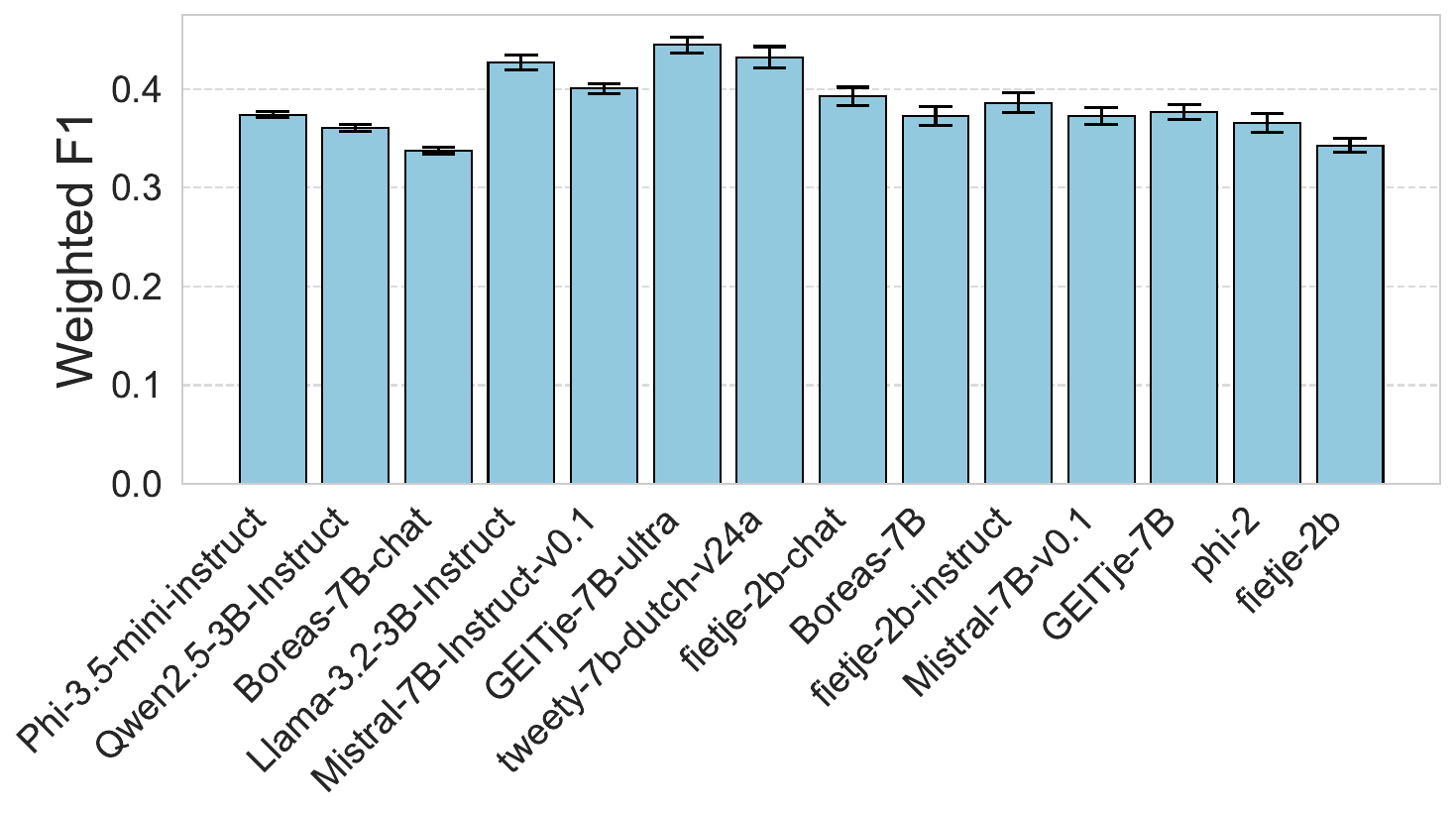}
        \caption{XLWIC}
        \label{fig:barplots-xlwic}
    \end{subfigure}

    \caption{Results per benchmark}
    \label{fig:results}
\end{figure}

\paragraph{Fietje} Fietje's results provide several insights into its capabilities and limitations. First and perhaps most surprising, the base Fietje model sometimes underperforms compared to Phi 2, the model it was derived from, although the confidence intervals suggest these results are close enough to be considered a toss-up. However, Fietje also notably outperforms Phi 2 on reasoning tasks like ARC (reasoning; Fig.~\ref{fig:barplots-arc}) and MMLU (knowledge and understanding; Fig.~\ref{fig:barplots-global_mmlu}), indicating successful adaptation for knowledge-based tasks.

The instruct and chat versions of Fietje show substantial improvements over the base version. Fietje 2B Chat, in particular, performs remarkably well for its size, outperforming larger 7B models like GEITje and Tweety on multiple benchmarks, including ARC and MMLU. Specifically, Fietje Chat surpasses GEITje Ultra and Tweety in two out of five tasks. This demonstrates the efficacy of instruction tuning and chat-specific adaptations in enhancing model performance. Especially at the time of its release, when models like Boreas, Phi 3.5, Llama 3.2 and Qwen 2.5 did not exist yet, Fietje showed to clearly perform as the best model on Dutch in its weight category.

\paragraph{Performance of modern, small, multilingual models} The benchmark results highlight the exceptional performance of small, multilingual models like Qwen 2.5 and Phi 3.5. This improved performance on Dutch benchmarks illustrates a welcome trend: size is not the sole determinant of model performance but multilingual pretraining efforts are key. These models, both under 4 billion parameters, outperform several 7B models on benchmarks like MMLU and ARC. Indeed, looking at the corresponding Figures \ref{fig:barplots-global_mmlu} and \ref{fig:barplots-arc}, very similar tendencies between models can be observed, and the top three models are visibly a step ahead of the others. While Phi 3.5, Qwen 2.5 and Llama 3.2 all perform well, a clear gap exists between the first two and Llama 3.2, which performs worse than the others. This gap may be attributed to the training data: while it is unclear (but likely) whether Phi 3.5 and Qwen 2.5 trained on Dutch, the Llama 3.2 model card explicitly mentions that it is only trained on English, German, French, Italian, Portuguese, Hindi, Spanish and Thai -- no Dutch. Yet, intriguingly, its tokenizer has a lower fertility on Dutch data than the others.

\paragraph{Model performance consistency} A notable observation in the benchmark results is the consistent performance of the top 5 models across multiple tasks. Qwen 2.5, Phi 3.5, Boreas Chat, Llama 3.2, and Mistral 7B Instruct frequently occupy the leading positions in tasks such as Global MMLU, DBRD, and ARC. This consistency suggests that while these models differ in size and architecture, their training data and multilingual capabilities give them an advantage across diverse benchmarks. The odd-one-out in this respect is Mistral 7B Instruct, which was not explicitly trained on Dutch.

\paragraph{Performance on XLWIC}
However, even despite this consistency, one task catches the eye. The XLWIC benchmark, which tests semantic disambiguation, stands out as an anomaly compared to the other tasks (visualized in Figure~\ref{fig:barplots-xlwic}). Several models that perform well across tasks like ARC or MMLU, such as Boreas Chat and Qwen 2.5, perform poorly on XLWIC. For instance, Boreas Chat ranks worst on XLWIC despite being in the top 4 for other tasks. While it is tempting to argue that XLWIC is simply a task that requires intricate knowledge of the Dutch language, explaining why Dutch-adapted models such as GEITje 7B Ultra and Tweety perform so well, this does not hold: Mistral 7B Instruct ranks 4th on this task, even though it was not trained on Dutch.

\paragraph{Performance gaps between worst and best model} Performance gaps between models are particularly pronounced in the ARC (reasoning; Fig.~\ref{fig:barplots-arc}) and DBRD (sentiment analysis; Fig.~\ref{fig:barplots-dbrd}) benchmarks. For ARC, the top-performing model, Qwen 2.5, achieves a score of 66.97\%, while the worst-performing model, Phi 2, scores only 18.07\%, reflecting a significant disparity in reasoning skills. A similar trend is visible in DBRD, where Boreas Chat achieves the highest score (94.38\%), incredibly close to the state-of-the-art for the benchmark of 95.14\% F1 \cite{delobelle-etal-2020-robbert}, which required finetuning a Dutch encoder model on the task. On the other end of the spectrum, Phi 2 only reaches 51.45\%. In the case of Boreas Chat, this may not be a surprise since it was trained on an undisclosed set of Dutch literature. In fact, out of all Mistral derivatives, Boreas is the best-ranking model, which is likely due to its varied Dutch-English dataset. A curious exception is its poor performance in the disambiguation task WIC, where it scores worst of all models (Fig.~
\ref{fig:barplots-xlwic}).

\paragraph{Mistral 7B Instruct's surprising performance} The Mistral 7B Instruct model demonstrates strong overall performance, with the notable exception of the Dutch CoLA benchmark (Fig.~\ref{fig:barplots-dutch_cola}), where it ranks last among all models. Despite being older and not being trained explicitly on Dutch according to the model information, it outperforms several Dutch-specific models. This indicates that Mistral's general training data may still provide a reasonable understanding of Dutch, although grammatical nuances remain a challenge. Of course it is also possible that the brief model description neglects to mention multilinguality when in fact the training data was multilingual, however the higher (worse) tokenizer fertility does not point in that direction.

\section{Discussion}

In this work, I have presented Fietje, a continue-pretrained version of Phi 2. The creation of Fietje is marked by an emphasis on reproducibility and transparency. Fietje's training and the evaluations presented here are completely reproducible. Similarly, the data is publicly available.

Fietje proved to be a competitive model for its size when it was launched,  exhibiting results comparable to or exceeding GEITje 7B Ultra on benchmarks such as MMLU, Dutch CoLA, and ARC. Yet it was surpassed relatively quickly in the months after its release. The benchmark results presented in this paper illustrate that developments in LLM move quickly and that recent endeavors by large actors (like Meta AI and Microsoft) are thankfully incorporating more multilingual material, even in their smaller models. At the same time, the results also show the promise of language adaptation through language-specific changes of a tokenizer \cite{remy2024transtokenization} and continue-pretraining on high-quality data (Boreas). A combination of the two would likely lead to a model of very high quality.

Not only is high quality of importance, but continue-pretraining on both English and Dutch as done in Boreas may continue to achieve better alignment between the initial model and the new target language. In addition, it is clear that Boreas' data mix has a significant impact on its performance: it clearly outperforms other Mistral v0.1 derivatives in most benchmarks thanks to its undisclosed dataset and/or mixing in English.

Another aspect that deserves more attention is the role of post-training, the stage after pre-training (instruction tuning and potentially alignment). Recent models such as Llama 3 \cite{grattafiori2024llama3} and Boreas highlight the significance of extensive post-training procedures, which can significantly improve performance across benchmarks. Yet, for Dutch, public datasets for any form of post-training are sorely scarce.

Focusing on the benchmarks themselves, it is worth emphasizing that machine-translated benchmarks like ARC and Global MMLU, may inadvertently favor some models due to ``translationese effects''. These artifacts of translation can introduce stylistic biases that benefit models trained on similar data. For this reason, other machine-translated benchmarks were not included in this paper (see Sec.~\ref{sec:limitation-eval}. We should therefore aim to create new, human-created (or at least human-corrected) benchmarks.

In addition, quantitative benchmarks offer valuable insights, but it is important to distinguish between language understanding and language production in large language models. As discussed by \citeasnoun{vanroy2023languageresourcesdutchlarge} and \citeasnoun{vanroy2024geitje7bultraconversational}, models trained predominantly on English data may perform well on classification tasks in other languages but may struggle with fluent language generation. For example, non-English benchmarks often require models to produce simple labels such as `positive' or `negative', or to select multiple-choice answers (`A', `B', etc.). Such tasks differs fundamentally from generating fluent, grammatically correct Dutch text. While such models may serve as effective zero-shot classifiers, a robust conversational assistant must excel in both comprehension and language production. Unfortunately, current benchmarks lack comprehensive metrics for evaluating fluency, highlighting a gap that future research must address. It is therefore always recommended to a select an LLM based on a user's needs and not merely by the reported numbers on benchmarks.

In conclusion, this paper has introduced Fietje and its Instruct and Chat derivatives, as well as all materials needed to fully and openly reproduce their creation and evaluation. While it performs well for its size, especially at the time of publication, it is clear that the field has not stood still in the months following its release. This paper has also critically examined a number of others LLMs, both specifically tuned for Dutch or not, and found that model size has less of an impact than a model's release date, since smaller, more recent models are outperforming models twice their size. This tendency indicates a hopeful way forward where small models are also multilingually trained, making language technology more accessible for Dutch language users.

\section{Limitations}

While this paper is mostly a system description paper, it also comes with its set of limitations, both in the way the model was created and how it is evaluated.

\subsection{Model} 

While Phi 2 was the best model of its weight at the time of training, this quickly changed. Working on language adaptation of LLMs, one always ``lags behind'': once you are done with the language adaptation of the model (collecting and filtering data, setting up training, acquiring compute, training), a new model has already been released that is potentially better suited to start from. Furthermore, Fietje was only trained on Dutch CulturaX and Wikipedia, but current approaches tend to incorporate a lot more math and programming code, e.g. Llama 3 was trained on 25\% mathematical and reasoning data and 17\% code \cite{grattafiori2024llama3}. In addition, in the ``post-training'' phase (supervised finetuning), it would appear that models such as Boreas Chat also hint towards the importance of high-quality, high-volume instruction tuning. While Fietje Instruct was trained on large, general knowledge instruction sets, this can definitely be improved. Since more such datasets are not publicly available for Dutch, there are many more available for English. So continue-pretraining on both Dutch and English datasets in both the pretraining and post-training phase, like Boreas, is an appealing avenue. Massive pretrain corpora for Dutch of high quality are also a topic of need and interest, which has attracted wider attention, as is evident from the release of the Dutch portion of FineWeb 2 \cite{penedo2024fineweb-2}.

\subsection{Evaluation}\label{sec:limitation-eval}

Upon Fietje's release in early 2024, it was relatively straightforward to evaluate and interpret the results. With the GEITje models as one of the few, if not the only, LLMs that were fluent in Dutch, not too large ($<10$B parameters), and performant in benchmarks, they were the key models to compare with. However, since then many model builders have put more emphasis on multilingualism.

Because of the heterogeneous nature of the models in terms of pretraining data, training mechanism, language focus, and small size differences, it is complex to derive convincing tendencies in the benchmark results. There are so many unknowns about most of these models that it is not evident to pinpoint differences and similarities that may explain performance. Even more so, the benchmark results on the words-in-context benchmark (WIC) illustrate how seemingly top-of-the-line models that perform consistently in the top-3 in other benchmarks, may still falter in specific benchmarks of a different nature. To this end, it may be worthwhile to expand the evaluation scope even further to benchmarks that were not included in this paper, such as the machine-translated HellaSwag \cite{zellers-etal-2019-hellaswag} benchmark (given a cut-off sentence, pick the best sentence continuation), a part-of-speech benchmark like CoNLL \cite{tjong-kim-sang-2002-introduction}, or a question-answering benchmark like the machine-translated SQuAD \cite{rajpurkar-etal-2018-know}. However, I did not include these because of the computational cost of running benchmarks for all models (and each benchmark five times to compute confidence intervals), but also because I wanted to limit the number of machine-translated benchmarks in favor of benchmarks that included ``natural Dutch'' such as DBRD, Dutch CoLA and XLWIC-NL. CoNLL has the additional issue that the expected output response must be in valid JSON. While tools like Outlines can ``force'' JSON output, such a requirement still imposes negative biases to models that were not trained on JSON as much -- which leads to the question whether the benchmark is actually measuring question-answering skills or JSON-production prowess. In conclusion, evaluating language models is hard, and for Dutch specifically we are in need of higher-quality, Dutch-native benchmarks. Not only because of the Dutch fluency, but also to better take into account the culture of the people that use Dutch as their main language.

An additional limitation of this paper is that the benchmarks are only tested in a zero-shot setting. Few-shot benchmarks used to be needed to improve the chances that the model would actually return the expected format or label, but with constrained decoding that is no longer as important -- though it is of course more than likely that benchmark results would improve with few-shot examples. Secondly, a potential shortcoming is that only one prompt was used for each task. It is possible that one task formulation favors one model better than the other, but it is not feasible to do an extensive search for the most optimal prompt for each model for each task. Therefore the current prompt templates (App.~\ref{app:benchmark-template}) were selected based on a consistent and natural formulation across benchmarks.

\section*{Acknowledgments}

Fietje and derivatives were trained on the VSC Tier-1 cluster of the Flemish SuperComputer Center (Vlaams Supercomputer Centrum; \url{https://www.vscentrum.be/}) on grant number 2023\_071.

\bibliographystyle{clin} 
\bibliography{references}  

\appendix

\section{Bad word list}\label{app:bad-words}

\textbf{\textcolor{red}{This word list contains offensive words!}} Documents containing any of these words were not included in the final training dataset.

Note that this is a strict list, included ambiguous words such as ``zak'' and ``uilskuiken'' that also have non-offensive readings. Nevertheless, for the sake of continue-pretraining (rather than from-scratch pretraining) on high quality, it was decided to have a wide coverage of potential bad words. In future versions, this list will be refined and ``bad documents'' will be removed with the addition of different metrics rather than only a word list, e.g. by automatic classification or perplexity measures.

\begin{lstlisting}[style=python]
BAD_PHRASES_DOC_LEVEL = {
    # https://en.wikipedia.org/wiki/Dutch_profanity
    "achterlijk",
    "debiel",
    "downie",
    "idioot",
    "kankerlijer",
    "klere",
    "kolere",
    "minkukel",
    "pestkop",
    "pleuris",
    "pleuritis",
    "teringlijer",
    "tyfuslijer",
    "gadver",
    "getver",
    "godver",
    "godskolere",
    "godverork",
    "graftak",
    "kopvod",
    "verdomme",
    "anaalgeneraal",
    "bitch",
    "dikzak",
    "flikker",
    "fok",
    "fuck",
    "hoer",
    "klootzak",
    "klote",
    "kreng",
    "kringspiermusketier",
    "kut",
    "lamzak",
    "lul",
    "manwijf",
    "matennaai",
    "neuken",
    "neuker",
    "ouwehoer",
    "reet",
    "reetkever",
    "reetridder",
    "rotzak",
    "schijt",
    "shit",
    "slet",
    "slijmbal",
    "slons",
    "sodemieter",
    "stoephoer",
    "swaffel",
    "teef",
    "trut",
    "tut",
    "zak",
    "uilskuiken",
    "zeik",
    "bamivreter",
    "bosneger",
    "neger",
    "fransoos",
    "geitenneuker",
    "kaaskop",
    "kakker",
    "koelie",
    "lijp",
    "medelander",
    "mocro",
    "mof",
    "nikker",
    "poepchinees",
    "roetmop",
    "spaghettivreter",
    "loempiavouwer",
    "spanjool",
    "spleetoog",
    "tatta",
    "tokkie",
    "zandneger",
    "zwartzak",
    "halvezool",
    "kenau",
    "klootviool",
    "knuppel",
    "koekert",
    "koekwaus",
    "oelewapper",
    "smeerlap",
    "sukkel",
    "sul",
    "wappie",
    "wijf",
    "zooi",
    # xxx (a.o. https://gitlab.com/yhavinga/c4nlpreproc/-/blob/master/clean/badwords_ennl.py)
    "xxx",
    "anal",
    "blowjob",
    "buttplug",
    "cock",
    "cunt",
    "geil",
    "sex",  # Standaardnederlands = seks, maybe we catch some porn or socialmedia sites with this misspelling
    "porn",
    # extra
    "nigger",
    "nigga",
    "hoerig",
    "klojo",
}    
\end{lstlisting}

\section{Training configuration}

In all cases, training was run with the alignment-handbook codebase \cite{tunstall2024alignment}. For the purpose of continued pretraining, I pushed a new task to the alignment-handbook that is now available for anyone who wants to use the codebase for further pretraining. Training runs were run on the Flemish Supercomputer (VSC).

Model creation is fully reproducible thanks to open data and the available configuration files. Below the alignment-handbook config files are given. They can also be found on GitHub.\footnote{\url{https://github.com/BramVanroy/fietje-2/tree/main/training}}

\subsection{Fietje 2B (base model)}\label{app:training-base}

\begin{lstlisting}[language=yaml]
# Model arguments
model_name_or_path: microsoft/phi-2
model_revision: main
torch_dtype: bfloat16
use_flash_attention_2: true
bf16: true
tf32: true

# Training arguments
learning_rate: 9.0e-05
adam_beta1: 0.9
adam_beta2: 0.98
adam_epsilon: 1.0e-7
weight_decay: 0.1
logging_steps: 1
logging_strategy: steps
lr_scheduler_type: linear
max_seq_length: 2048

per_device_train_batch_size: 40
per_device_eval_batch_size: 40

gradient_accumulation_steps: 3
gradient_checkpointing: true
gradient_checkpointing_kwargs:
  use_reentrant: False

# Data training arguments
dataset_mixer:
  /dodrio/scratch/projects/2023_071/alignment-handbook/data/fietje-2b-cpt-prep: 1.0
dataset_splits:
  - train
  - test
preprocessing_num_workers: 8
num_train_epochs: 1.0
remove_unused_columns: true
push_to_hub: true
report_to:
- wandb
log_level: info

# To do or not to do
do_train: True
do_eval: True
seed: 42

# Storing
output_dir: /dodrio/scratch/projects/2023_071/alignment-handbook/data/fietje-2b
overwrite_output_dir: true
save_total_limit: 6
hub_model_id: fietje-2b
hub_private_repo: true
hub_strategy: all_checkpoints

# Strategies
evaluation_strategy: "steps"
eval_steps: 900
save_strategy: "steps"
save_steps: 900
warmup_steps: 0

\end{lstlisting}

\subsection{Fietje 2B Instruct (instruction model)}\label{app:training-instruct}

\begin{lstlisting}[language=yaml]
# Model arguments
model_name_or_path: BramVanroy/fietje-2b
model_revision: main
torch_dtype: bfloat16
use_flash_attention_2: true
bf16: true
tf32: true

# Training arguments
learning_rate: 6.0e-05
adam_beta1: 0.9
adam_beta2: 0.98
adam_epsilon: 1.0e-7
weight_decay: 0.1
logging_steps: 1
logging_strategy: steps
lr_scheduler_type: cosine
max_seq_length: 2048

per_device_train_batch_size: 42
per_device_eval_batch_size: 36

gradient_accumulation_steps: 1
gradient_checkpointing: true
gradient_checkpointing_kwargs:
  use_reentrant: False

# Data training arguments
chat_template: "{% for message in messages %}{{'<|im_start|>' + message['role'] + '\n' + message['content']}}{% if (loop.last and add_generation_prompt) or not loop.last %}{{ '<|im_end|>' + '\n'}}{% endif %}{% endfor %}{% if add_generation_prompt and messages[-1]['role'] != 'assistant' %}{{ '<|im_start|>assistant\n' }}{% endif %}"
dataset_mixer:
  BramVanroy/ultrachat_200k_dutch: 1.0
  BramVanroy/no_robots_dutch: 1.0
  BramVanroy/belebele_dutch: 1.0
dataset_configs:
  - default
  - default
  - sft
dataset_splits:
  - train_sft
  - test_sft
preprocessing_num_workers: 8
num_train_epochs: 3.0
remove_unused_columns: true
push_to_hub: true
report_to:
- wandb
log_level: info

# To do or not to do
do_train: True
do_eval: True
seed: 42

# Storing
output_dir: /dodrio/scratch/projects/2023_071/alignment-handbook/data/fietje-2b-sft
overwrite_output_dir: true
hub_model_id: BramVanroy/fietje-2b-sft
hub_private_repo: true
hub_strategy: all_checkpoints
save_total_limit: 6

# Strategies
evaluation_strategy: "epoch"
save_strategy: "epoch"
warmup_ratio: 0.1
\end{lstlisting}

\subsection{Fietje 2B Chat (preference model)}\label{app:training-chat}

\begin{lstlisting}[language=yaml]
# Model arguments
model_name_or_path: BramVanroy/fietje-2b-sft
model_revision: main
torch_dtype: bfloat16
use_flash_attention_2: true
bf16: true
tf32: true

# Training arguments
learning_rate: 2.0e-06
adam_beta1: 0.9
adam_beta2: 0.98
adam_epsilon: 1.0e-7
weight_decay: 0.1
logging_steps: 1
logging_strategy: steps
lr_scheduler_type: cosine
max_length: 2048
max_prompt_length: 1280

# DPO
beta: 0.2

per_device_train_batch_size: 8
per_device_eval_batch_size: 4

gradient_accumulation_steps: 2
gradient_checkpointing: true
gradient_checkpointing_kwargs:
  use_reentrant: False

# Data training arguments
dataset_mixer:
  BramVanroy/ultra_feedback_dutch_cleaned: 1.0
  BramVanroy/orca_dpo_pairs_dutch_cleaned: 1.0
dataset_configs:
  - dpo_hq
  - dpo_all
dataset_splits:
  - train_prefs
  - test_prefs
preprocessing_num_workers: 8
num_train_epochs: 1.0
remove_unused_columns: true
push_to_hub: true
report_to:
- wandb
log_level: info

# To do or not to do
do_train: True
do_eval: True
seed: 42

# Storing
output_dir: /dodrio/scratch/projects/2023_071/alignment-handbook/data/fietje-2b-dpo
overwrite_output_dir: true
hub_model_id: BramVanroy/fietje-2b-dpo
hub_private_repo: true
hub_strategy: all_checkpoints
save_total_limit: 6

# Strategies
evaluation_strategy: "epoch"
save_strategy: "epoch"
warmup_ratio: 0.1
\end{lstlisting}

\section{Benchmark templates}
\label{app:benchmark-template}

These prompt templates are also available in the config files of the benchmarks at \url{https://github.com/BramVanroy/clin34-benchmarks/tree/main/configs}.

\subsection{ARC}

For models without a chat template (``base'' models), the text ``Het antwoord is '' is added to the end of the prompt.

\begin{lstlisting}[language=jinja]
{{- instruction }}

{% set options = [
('A', option_a),
('B', option_b),
('C', option_c),
('D', option_d)
] -%}
{%- set available_options = options | selectattr('1', 'defined') | rejectattr('1', 'none') | list -%}
{%- if available_options -%}
Antwoordopties:
{%- for letter, option in available_options %}
{{ letter }}. {{ option }}
{%- endfor %}

Antwoord met {% for i in range(available_options|length) -%}
'{{ available_options[i][0] }}'{% if i + 2 == available_options|length %} of {% elif i + 1 < available_options|length %}, {% endif %}
{%- endfor -%}.
{%- endif -%}
\end{lstlisting}

\subsection{DBRD}

For models without a chat template (``base'' models), the text ``Het sentiment is '' is added to the end of the prompt.

\begin{lstlisting}[language=jinja]
Is het sentiment in de volgende Nederlandstalige boekrecensie positief of negatief?

Boekrecensie: {{ text }}

Antwoord met 'positief' of 'negatief'.
\end{lstlisting}

\subsection{Dutch CoLA}

For models without a chat template (``base'' models), the text ``De tekst is '' is added to the end of the prompt.

\begin{lstlisting}[language=jinja]
Is de volgende tekst grammaticaal (correct Nederlands) of ongrammaticaal (onjuist Nederlands)?

Tekst: {{ Sentence }}

Antwoord met 'grammaticaal' of 'ongrammaticaal'.
\end{lstlisting}

\subsection{Global MMLU}

For models without a chat template (``base'' models), the text ``Het antwoord is '' is added to the end of the prompt.

\begin{lstlisting}[language=jinja]
{{- question }}

{% set options = [
('A', option_a),
('B', option_b),
('C', option_c),
('D', option_d)
] -%}
{%- set available_options = options | selectattr('1', 'defined') | rejectattr('1', 'none') | list -%}
{%- if available_options -%}
Antwoordopties:
{%- for letter, option in available_options %}
{{ letter }}. {{ option }}
{%- endfor %}

Antwoord met {% for i in range(available_options|length) -%}
'{{ available_options[i][0] }}'{% if i + 2 == available_options|length %} of {% elif i + 1 < available_options|length %}, {% endif %}
{%- endfor -%}.
{%- endif -%}
\end{lstlisting}

\subsection{XLWIC}

For models without a chat template (``base'' models), the text ``De betekenis van `\{\{ target\_word \}\}' is '' is added to the end of the prompt.

\begin{lstlisting}[language=jinja]
Is de betekenis van '{{ target_word }}' in de volgende zinnen identiek of verschillend?

Zin 1: {{ example_1 }}
Zin 2: {{ example_2 }}

Antwoord met 'identiek' of 'verschillend'.
\end{lstlisting}

\section{Benchmark visualizations}

\subsection{Model performance vs. model size (per task)}
\label{app:size-v-perf}

\begin{figure}[!ht]
    \centering
    
    \begin{subfigure}[b]{0.45\textwidth}
        \centering
        \includegraphics[width=\textwidth]{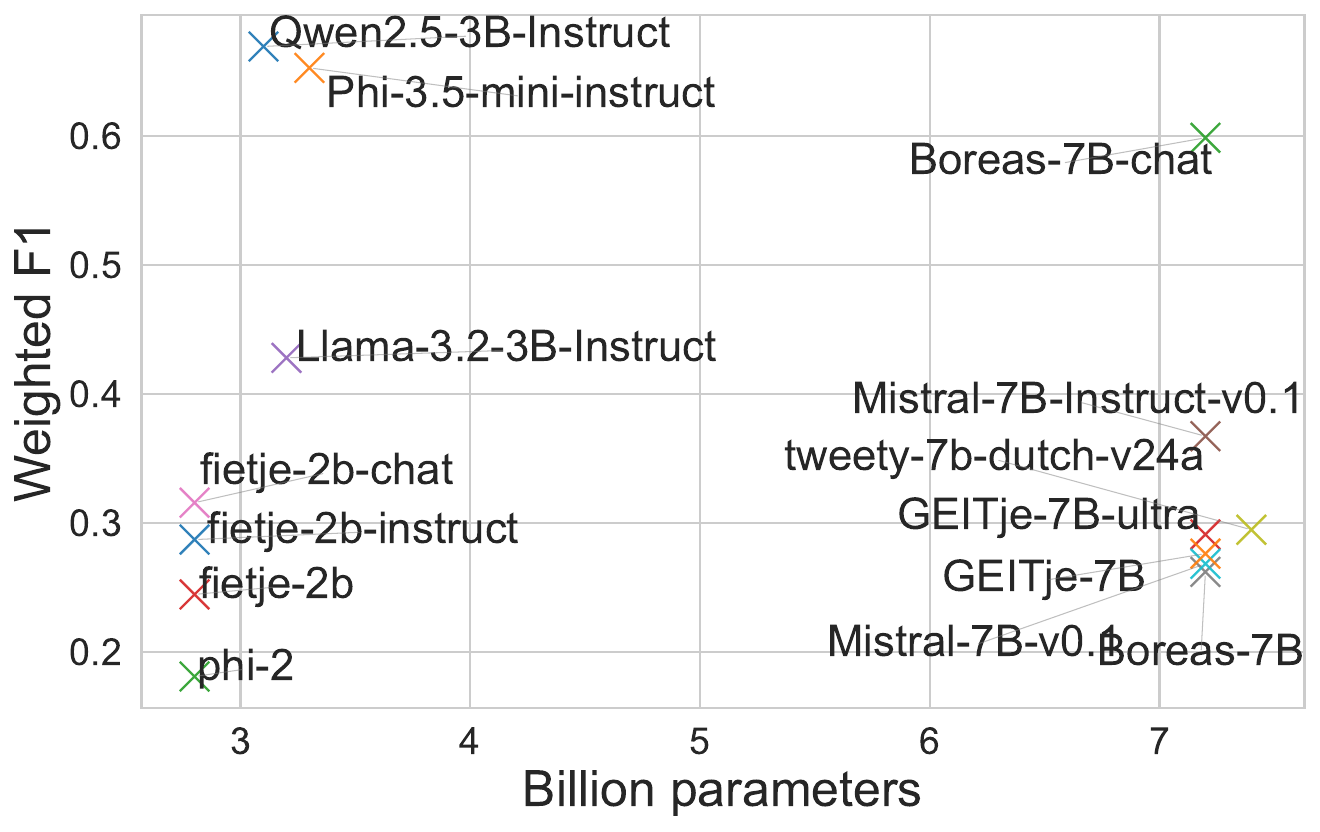}
        \caption{ARC}
        \label{fig:size-arc}
    \end{subfigure}
    \hfill
    \begin{subfigure}[b]{0.45\textwidth}
        \centering
        \includegraphics[width=\textwidth]{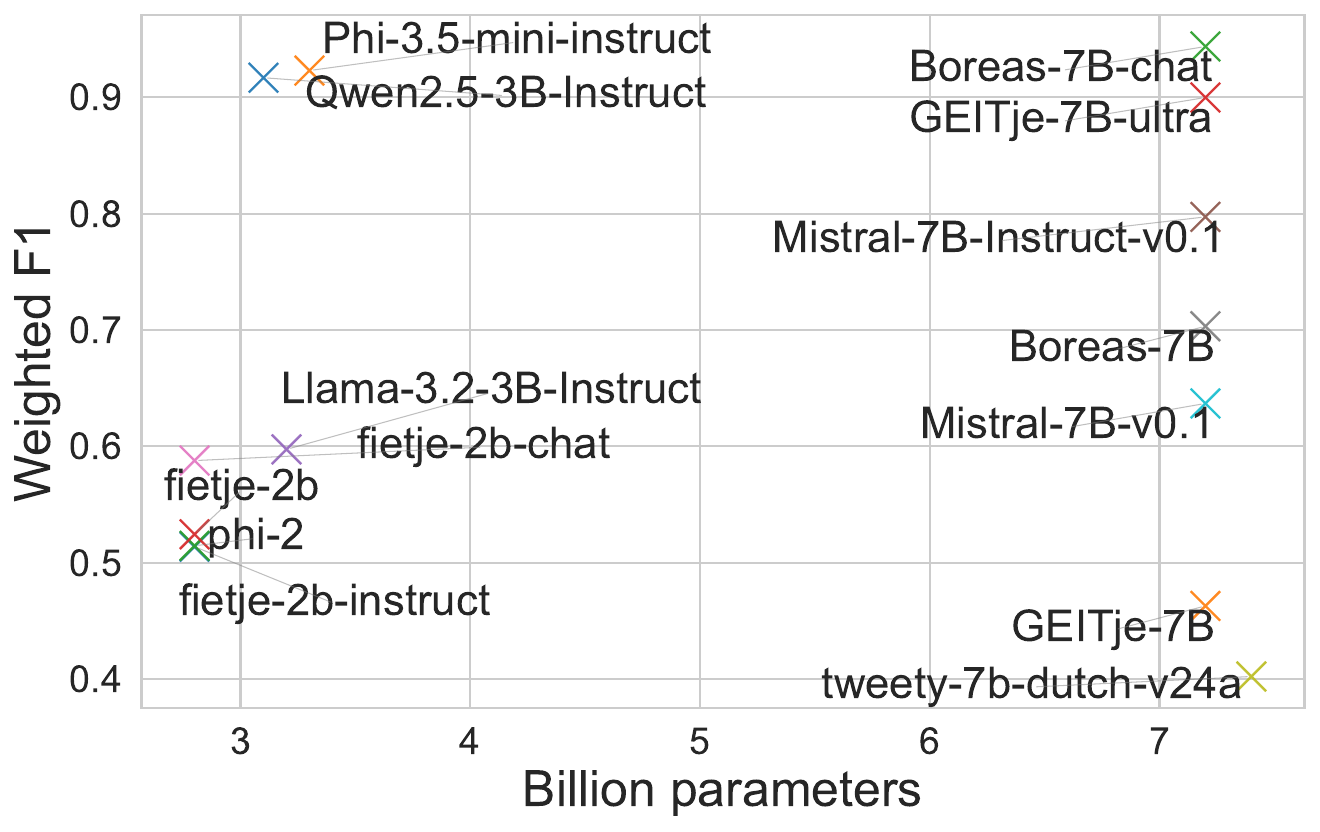}
        \caption{DBRD}
        \label{fig:size-dbrd}
    \end{subfigure}
    
    \vspace{0.5cm}

    \begin{subfigure}[b]{0.45\textwidth}
        \centering
        \includegraphics[width=\textwidth]{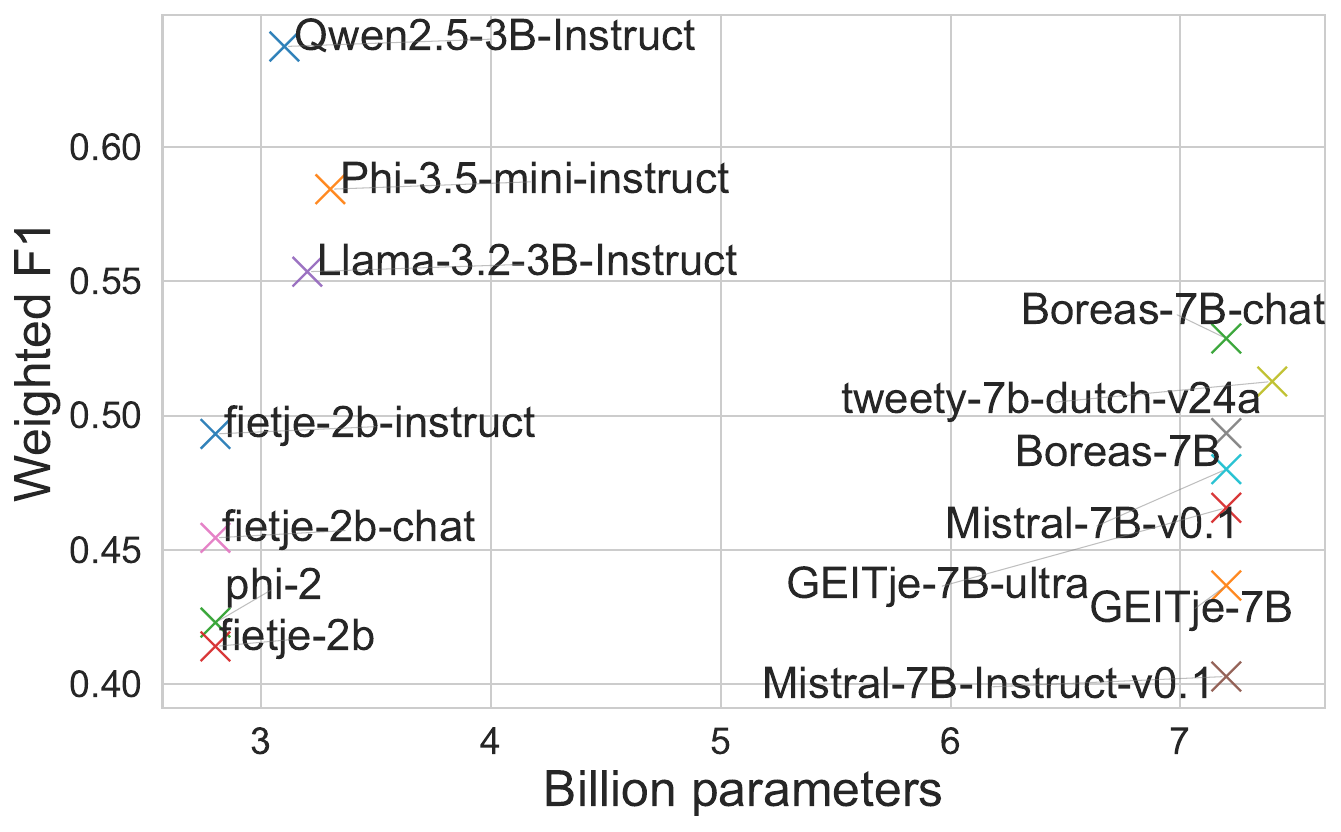}
        \caption{Dutch CoLA}
        \label{fig:size-dutch_cola}
    \end{subfigure}
    \hfill
    \begin{subfigure}[b]{0.45\textwidth}
        \centering
        \includegraphics[width=\textwidth]{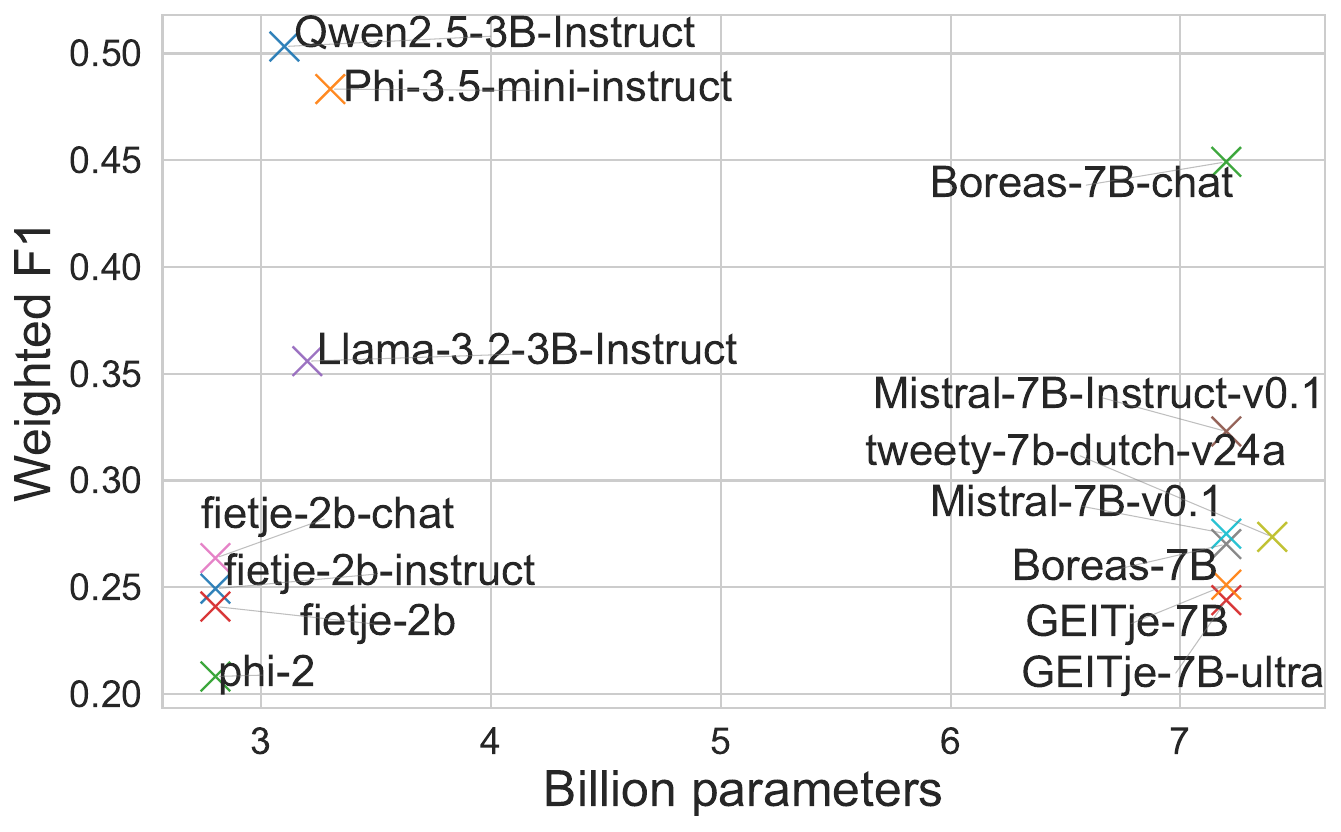}
        \caption{Global-MMLU}
        \label{fig:size-global_mmlu}
    \end{subfigure}

    \vspace{0.5cm}

    \begin{subfigure}[b]{0.45\textwidth}
        \centering
        \includegraphics[width=\textwidth]{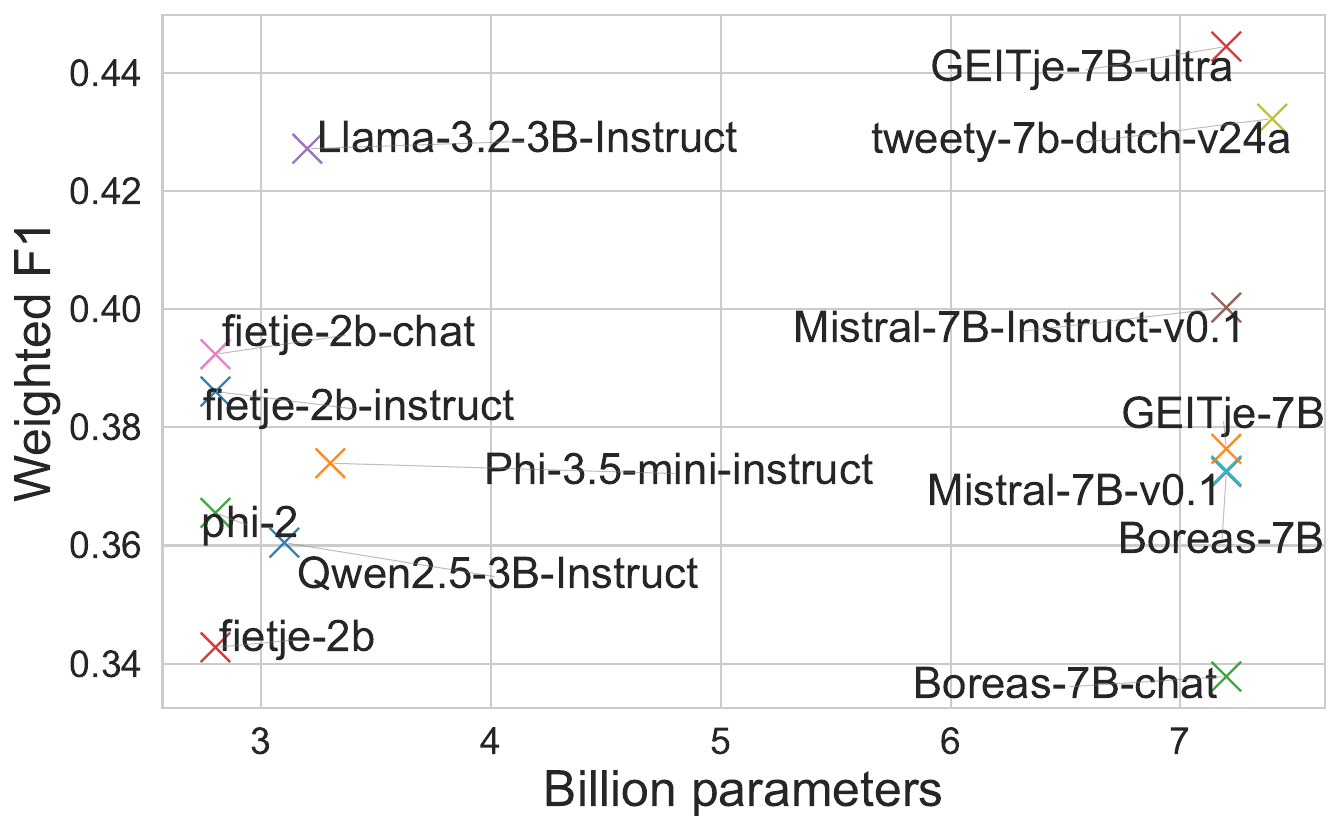}
        \caption{XLWIC}
        \label{fig:size-xlwic}
    \end{subfigure}

    \caption{Performance vs. size across all benchmarks}
    \label{fig:size-v-perf}
\end{figure}

\subsection{Model performance vs. model size (per task)}
\label{app:date-v-perf}

\begin{figure}[!ht]
    \centering

    \begin{subfigure}[b]{0.45\textwidth}
        \centering
        \includegraphics[width=\textwidth]{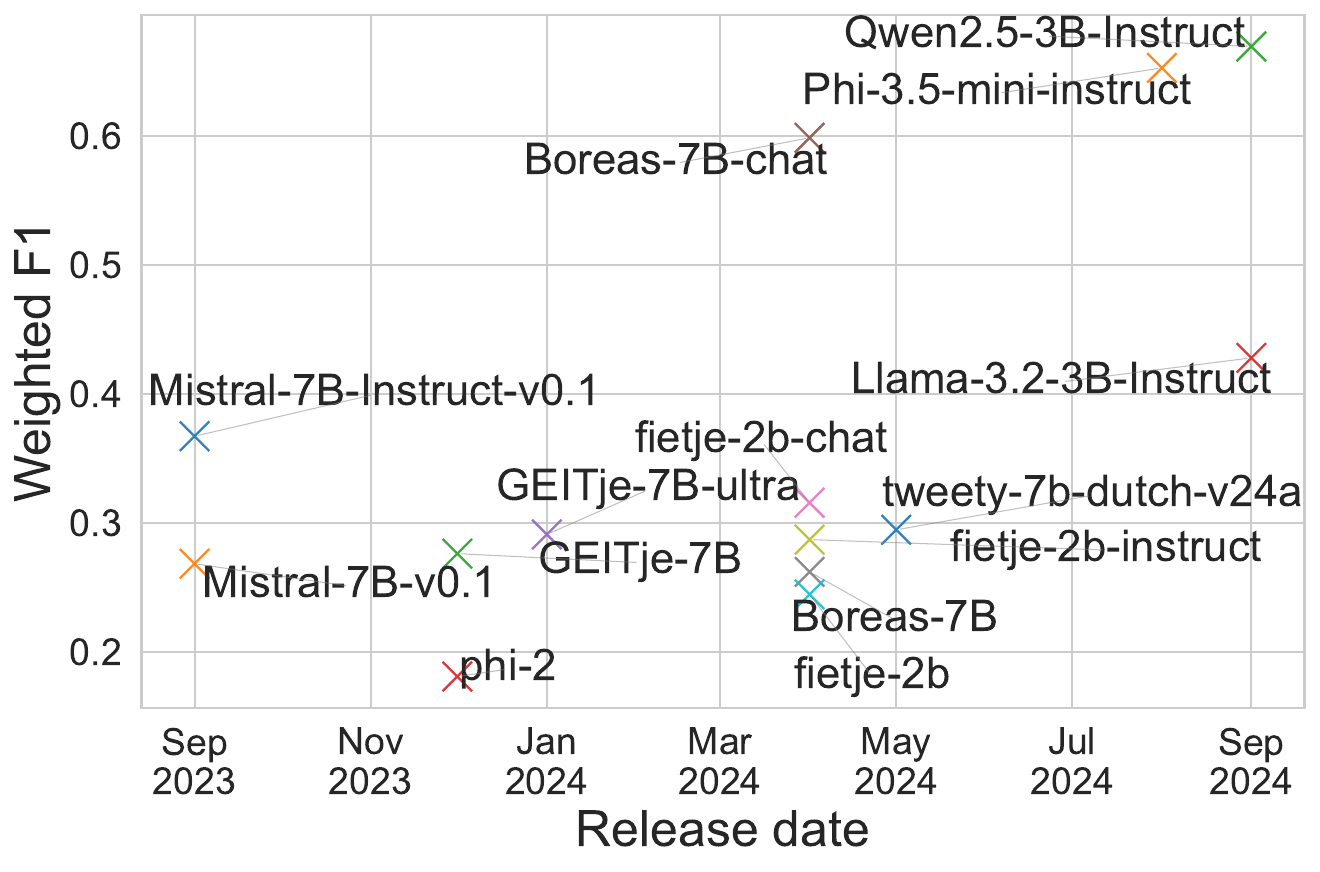}
        \caption{ARC}
        \label{fig:date-arc}
    \end{subfigure}
    \hfill
    \begin{subfigure}[b]{0.45\textwidth}
        \centering
        \includegraphics[width=\textwidth]{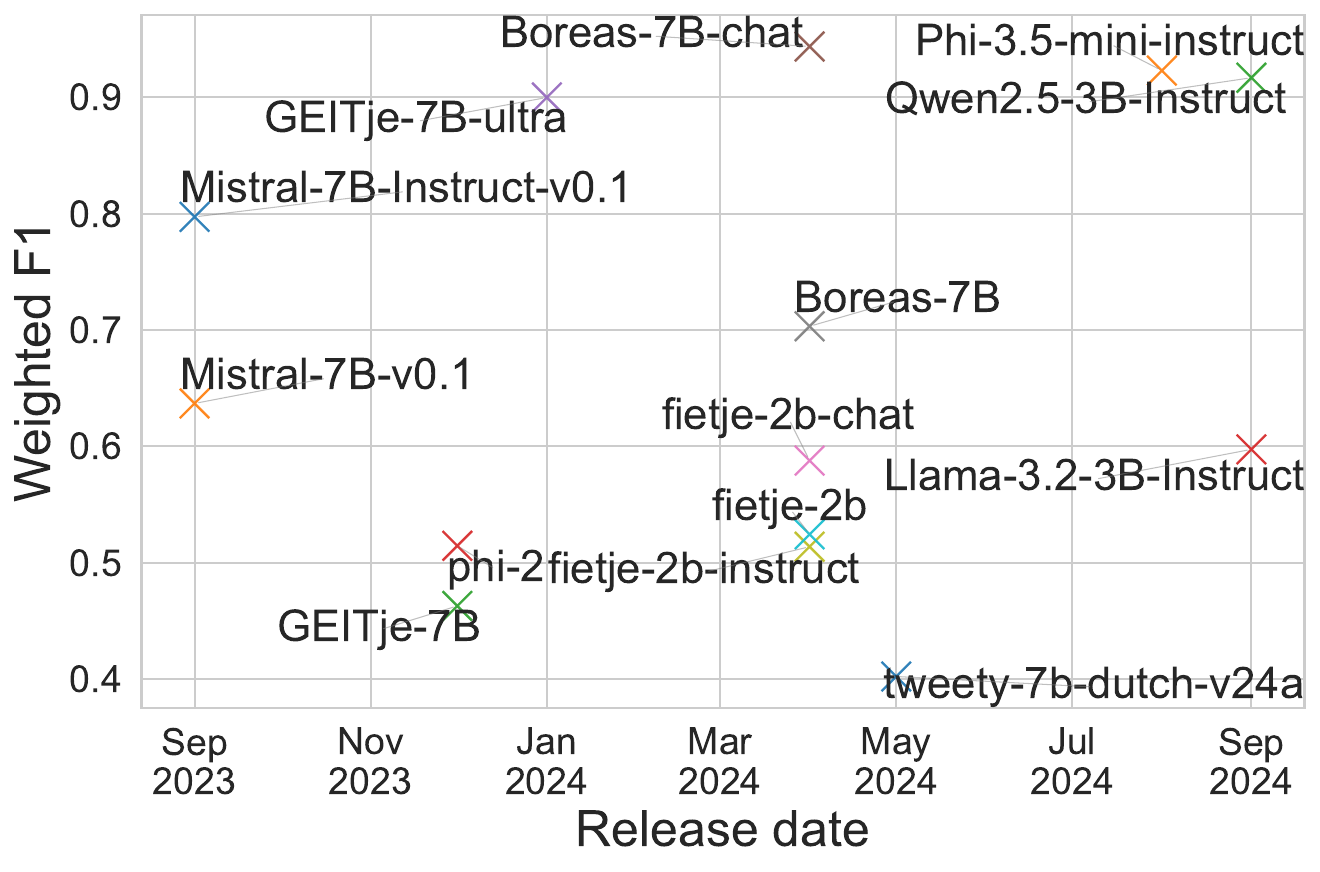}
        \caption{DBRD}
        \label{fig:date-dbrd}
    \end{subfigure}
    
    \vspace{0.5cm} 
    
    \begin{subfigure}[b]{0.45\textwidth}
        \centering
        \includegraphics[width=\textwidth]{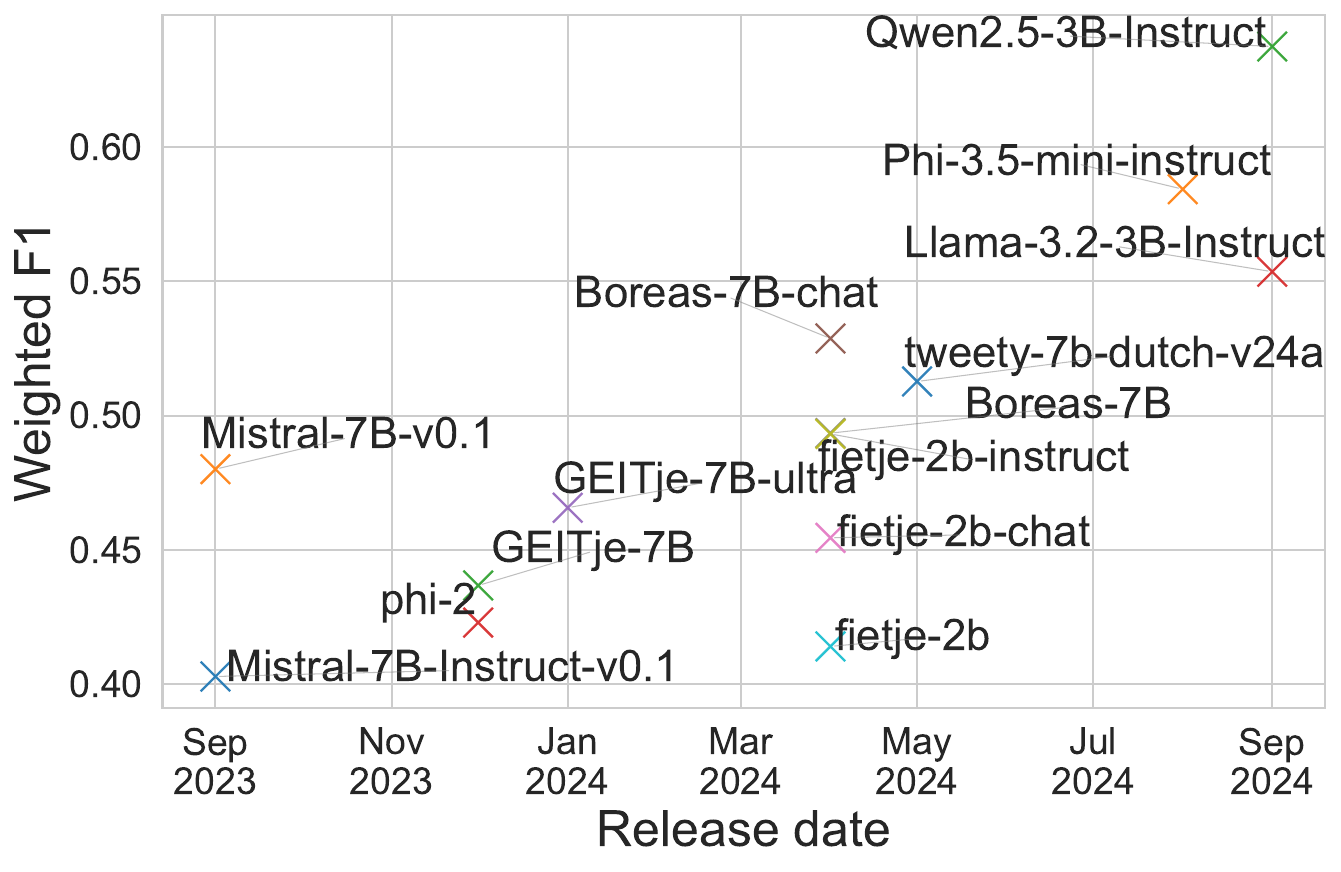}
        \caption{Dutch CoLA}
        \label{fig:date-dutch_cola}
    \end{subfigure}
    \hfill
    \begin{subfigure}[b]{0.45\textwidth}
        \centering
        \includegraphics[width=\textwidth]{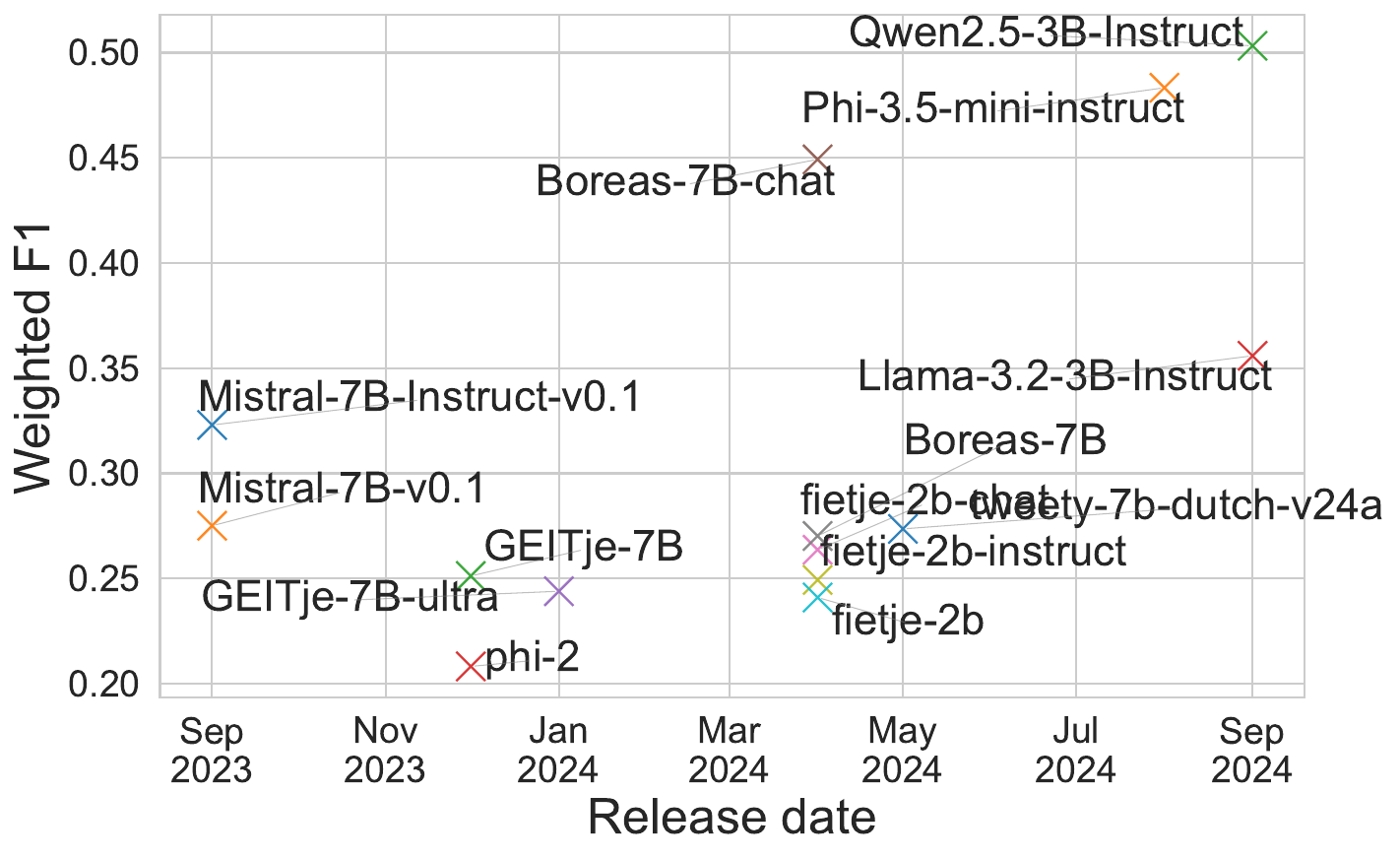}
        \caption{Global-MMLU}
        \label{fig:date-global_mmlu}
    \end{subfigure}

    \vspace{0.5cm} 

    \begin{subfigure}[b]{0.45\textwidth}
        \centering
        \includegraphics[width=\textwidth]{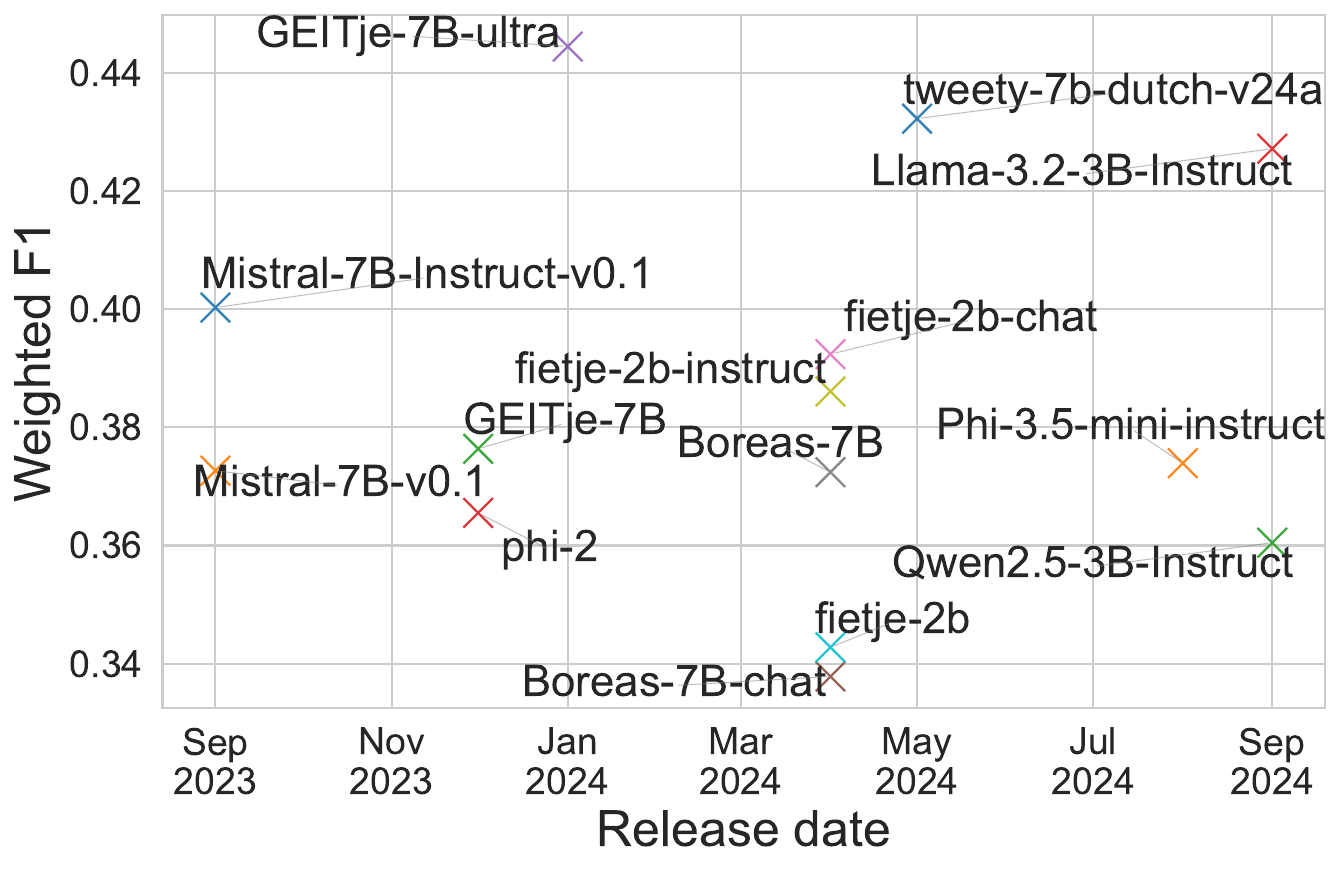}
        \caption{XLWIC}
        \label{fig:date-xlwic}
    \end{subfigure}

    \caption{Performance vs. release date across all benchmarks}
    \label{fig:date-v-perf}
\end{figure}

\end{document}